\documentclass{article}

\usepackage{xcolor}
\usepackage[final]{corl_2026} %
\let\oldaddcontentsline\addcontentsline
\renewcommand{\addcontentsline}[3]{}
\usepackage{amsmath,amssymb}
\usepackage{booktabs}
\usepackage{colortbl}
\usepackage{graphicx}
\usepackage{multirow}
\usepackage{xspace}
\usepackage{enumitem}
\usepackage{bm}
\usepackage{wrapfig}
\usepackage{hyperref}
\hypersetup{pdftitle={µ0: A Scalable 3D Interaction-Trace World Model},
            pdfauthor={Seungjae Lee, Yoonkyo Jung, Jusuk Lee, Jonghun Shin, Amir Hossein Shahidzadeh, Yao-Chih Lee, H. Jin Kim, Jia-Bin Huang, Furong Huang}}

\newcommand{\datapipe}{TraceExtract\xspace}
\newcommand{\algoname}{\ensuremath{\mu_0}\xspace}
\newcommand{\modelname}{\ensuremath{\mu_0}\xspace}

\newcommand{\topic}[1]{\vspace{-0.5mm}\noindent\textbf{#1}}
\newcommand{\ANDcompact}{%
  \end{tabular}\hfil\linebreak[4]\hfil%
  \begin{tabular}[t]{c}\bfseries\rule{0pt}{13pt}\ignorespaces%
}
\definecolor{projectorange}{HTML}{F77F00}

\newif\ifshowrev
\showrevfalse
\ifshowrev

\else

\fi

\title{\modelname: A Scalable 3D Interaction-Trace World Model}

\author{%
  Seungjae Lee\textsuperscript{1*},
  Yoonkyo Jung\textsuperscript{1*},
  Jusuk Lee\textsuperscript{2},
  Jonghun Shin\textsuperscript{2},\ANDcompact
  Amir Hossein Shahidzadeh\textsuperscript{1},
  Yao-Chih Lee\textsuperscript{1},
  H.~Jin Kim\textsuperscript{2},\ANDcompact
  Jia-Bin Huang\textsuperscript{1\textdagger},
  Furong Huang\textsuperscript{1,3\textdagger}\\
  \\
  \textsuperscript{1}University of Maryland, College Park\quad\textsuperscript{2}Seoul National University\quad\textsuperscript{3}All Purpose AI\\
  \textsuperscript{*}Equal contribution\quad\textsuperscript{\textdagger}Equal advising\\\\
  {\fontsize{11}{12}\selectfont
  Project Page: \href{https://mu0-wm.github.io/}
  {\textcolor{projectorange}{https://mu0-wm.github.io/}}
  }%
}

\begin{document}
\maketitle

\vspace{-2em}

\begin{abstract}
World models that capture how actions induce physical change enable scalable robot learning without reliance on embodiment-specific action labels. Pixel-space video models provide broad visual priors but expend model capacity on dense appearance reconstruction, while direct action models require embodiment-specific labels that hinder scalability. We present \modelname, a scalable world model based on 3D traces.
Rather than predicting dense pixels or directly modeling actions, \modelname forecasts smooth 3D trajectories for salient interaction points such as objects, tools, hands, and contact regions, yielding a compact, embodiment-agnostic motion interface. To enable training from diverse video sources, our \datapipe system automatically extracts 3D supervision by selecting keypoints, constructing globally aligned traces, and associating motion segments with hierarchical language captions. This \datapipe supervision pretrains \modelname by combining a pretrained vision-language backbone with a modular trace expert, which represents each query via B-spline control points and predicts future traces. Experiments show that \modelname outperforms baselines in both 2D and 3D trace prediction, including trace prediction models and tokenized VLM methods.
Because \modelname is frozen and reusable, it can be paired with action experts for downstream robot embodiments.
Despite action-free pretraining, the resulting trace-conditioned policies achieve performance competitive with VLA models pretrained with action supervision, such as $\pi_0$.
These results establish 3D traces as a scalable and transferable representation for cross-embodiment manipulation.
\end{abstract}


\begin{figure*}[!htbp]
    \centering
    \vspace{-0.5em}
    \includegraphics[width=1\textwidth]{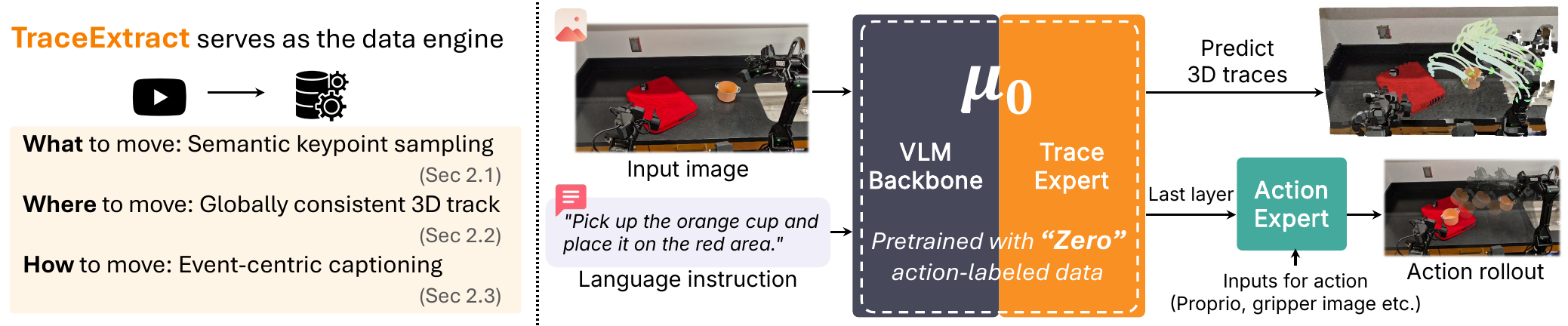}
    \vspace{-1.5em}
    \caption{\textbf{From videos to reusable action priors.} \datapipe extracts event-captioned 3D interaction traces from heterogeneous videos by selecting entity-centric keypoints, lifting them into globally aligned 3D, and pairing motion events with language. This supervision pretrains \modelname as a world model that predicts compact future trajectories for interaction points, instead of dense pixels or robot-specific actions. Once pretrained, the frozen \modelname can be reused with any downstream action expert, which consumes trace features to produce executable robot action chunks.
    }
    \label{fig:teaser}
    \vspace{-1em}
\end{figure*}

\section{Introduction}
Robot learning is constrained by a fundamental data paradox. 
On one hand, videos provide an abundant and scalable source of physical behavior data. 
On the other hand, the most useful kind of supervision for control, action-labeled robot data, is scarce, expensive, hardware-specific, and incompatible across embodiments.
World models offer a path around this bottleneck by learning from observation-rich video data and later grounding their predictions to specific robot embodiments~\citep{lin2026roboflow4d,gao2026dreamdojo,cho2026egoavflow,wang2026mvista,kim2026pri4r,wang2026eva}.
The key question is \textit{what such a model should predict}.
Pixel-space video generation is scalable but expends model capacity on dense appearance and background reconstruction, while often failing to capture the metric geometry, contact structure, and occlusion patterns required for manipulation~\citep{du2023learning,hu2025video,agarwal2025cosmos}.
Direct action prediction, as in Vision-Language-Action models, remains limited by the scarcity and embodiment specificity of labeled robot demonstrations.
We instead occupy the middle ground: 3D traces of semantic interaction points---object parts, tools, hands, and contact regions---which compactly describe what must move regardless of the robot used.

Recent motion-centric methods point in this direction through 2D flows~\citep{wen2024any,xu2024flowcrossdomainmanipulationinterface,nguyen2026pixel}, 3D flows~\citep{zhi20253dflowaction,huang2026pointworld,wang2026lamp,hung20263pointr, lee2026tracegen}, and object trajectories~\citep{bharadhwaj2024track2act}.
However, existing systems share three limitations: 
1) they under-sample small but task-critical regions such as tool tips and contact patches; 
2) they conflate object motion with camera motion by operating in local or 2D image-space coordinates; and 
3) they pair long demonstrations with episode-level captions rather than event-level intent.
These gaps motivate a trace world model that 
(i) selects where to measure motion, 
(ii) preserves global 3D structure, and 
(iii) binds local motion segments to language.
The closest prior work, TraceGen~\citep{lee2026tracegen}, predicts 3D traces on a fixed grid with episode-level captions and depth-conditioned input, and is therefore limited along all three axes; our system addresses these limitations.

We present \modelname, a query-conditioned 3D trace-space world model that serves as a reusable motion prior for downstream action experts (Fig.~\ref{fig:teaser}).
To supply training data at scale, we introduce \datapipe, a scalable data engine that converts heterogeneous human and robot videos into event-captioned trace supervision by 
(i) selecting semantic keypoints via DINOv2 entity clusters,
(ii) lifting them into globally aligned 3D, and 
(iii) captioning trace-driven motion events with hierarchical language---scaling trace curation by roughly $8\times$ over prior 3D trace datasets~\citep{lee2026tracegen}.
\modelname is built on a pretrained VLM backbone augmented with a permutation-equivariant Trace Expert, which forecasts flexible semantic keypoints as smooth B-spline traces using a semantic flow-matching objective.
After video-only pretraining, the frozen \modelname becomes a reusable motion prior: an Action Expert attending to its trace-denoising features, along with robot observations, proprioception, and language, outputs executable action chunks for any target embodiment.
On 2D/3D trace forecasting, \modelname outperforms prior trace prediction models and tokenized-VLM baselines. 
In 8 RoboCasa365 simulation \cite{nasiriany2026robocasa365} tasks and 3 real-world UR3 manipulation tasks, \algoname matches or exceeds action-labeled VLAs ($\pi_0$ \cite{black2025pi0}, $\pi_{0.5}$ \cite{intelligence2025pi_}), achieving 120--130\% of $\pi_0$'s and 70--115\% of $\pi_{0.5}$'s average success rates, despite using no action supervision during pretraining.

Our \textbf{main contributions} are:
\textbf{(1) \datapipe}, a scalable data engine that extracts event-captioned 3D trace supervision from heterogeneous manipulation videos via semantic keypoint selection, globally aligned 3D lifting, and hierarchical language captioning.
\textbf{(2) $\boldsymbol{\mu_0$}}, a query-conditioned 3D trace-space world model with a VLM backbone, permutation-equivariant Trace Expert, B-spline trace targets, and semantic flow-matching training.
\textbf{(3) Trace-conditioned action adaptation}, which freezes the pretrained \modelname and trains an action expert on top of its trace-denoising features, enabling action-free video pretraining to transfer to effective robot policies.

\section{\datapipe: A Scalable Cross-Embodiment Data Pipeline}
\label{sec:data_pipeline}
\topic{Measurement target.}
A trace-space world model must decide \emph{where} to measure motion. Dense pixels are redundant and background-heavy, while uniform grids waste queries on static surfaces and can miss small manipulated parts. Thus, we can predict \textbf{interaction-centric keypoints} on \textit{objects}, \textit{tools}, \textit{hands}, and \textit{contact regions}; their 3D motion captures what changes and what a robot should reproduce. These traces are embodiment agnostic---the same object motion can guide different robot morphologies---but only when they provide \textbf{(1) semantic selection}, so keypoints lie on task-relevant entities; \textbf{(2) consistent 3D tracking}, so identities survive camera motion and long horizons; and \textbf{(3) event-level language}, so local motion segments are paired with the right skill descriptions.

\topic{Pipeline overview.}
We introduce \datapipe, the data engine used to train \modelname. It treats trace extraction as interaction-centric supervision and, building on TraceGen~\citep{lee2026tracegen}, remedies fixed-grid, short-clip trace curation with \textbf{(1) task-relevant keypoints}, \textbf{(2) globally consistent 3D identities}, and \textbf{(3) language aligned to motion events}. These properties let \datapipe scale curation by producing \{observation, trace, language\} triplets for training \modelname (Sec.~\ref{sec:training}).

\begin{figure*}[t!]
    \centering
    \includegraphics[width=1\textwidth]{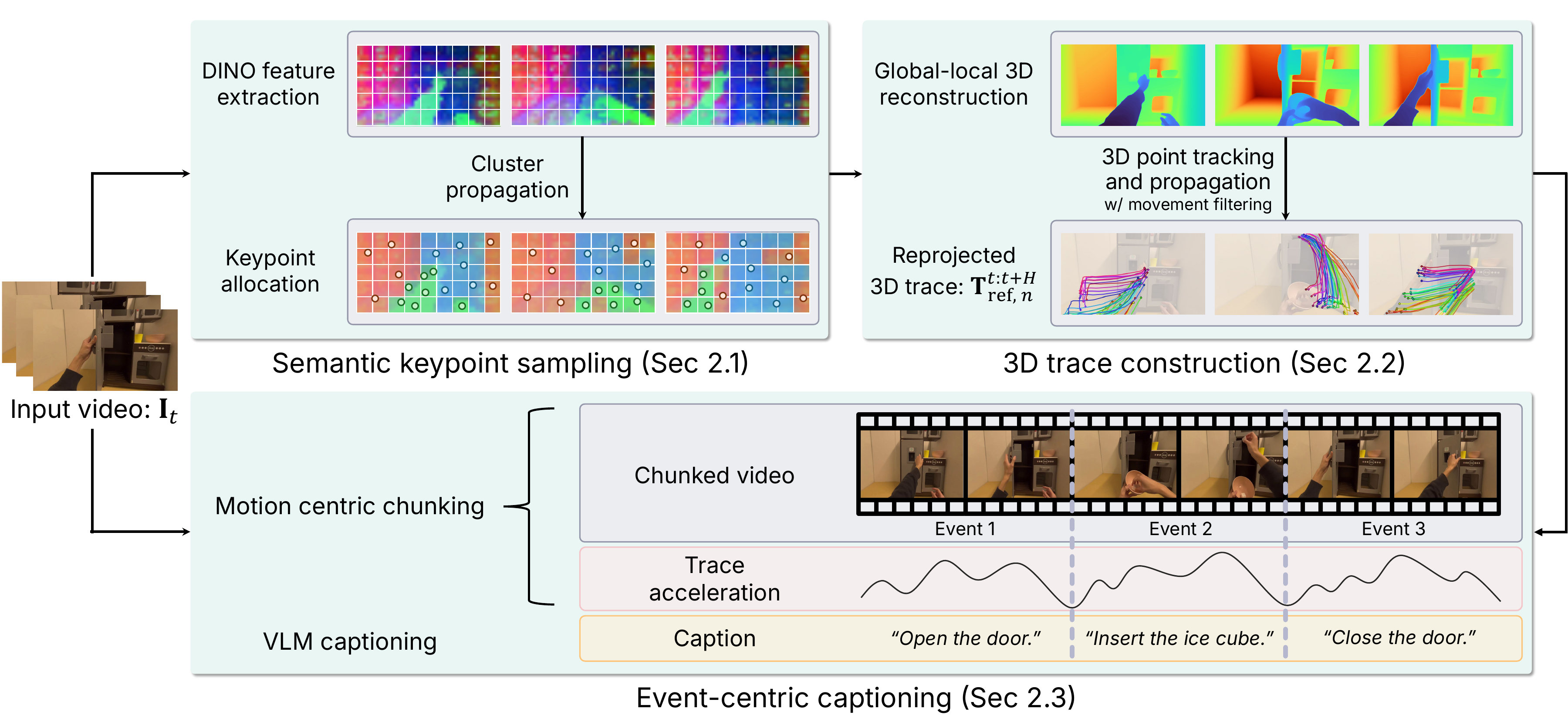}
    \vspace{-1em}
    \caption{\textbf{Overview of \datapipe.}
    From an uncurated human or robot manipulation video, \datapipe selects DINOv2 entity keypoints (Sec.~\ref{subsec:dino_keypoints}), tracks and lifts them into globally aligned 3D traces with chunk-wise reconstruction (Sec.~\ref{subsec:trace_construction}), and segments traces into motion-centric events for hierarchical VLM captioning (Sec.~\ref{sec:caption}), producing event-captioned 3D trace supervision for \modelname.}
    \label{fig:datapip}
    \vspace{-1em}
\end{figure*}

\subsection{Semantic Keypoint Sampling}
\label{subsec:dino_keypoints}
Prior fixed-grid trace extraction~\citep{lee2026tracegen} is simple but area-biased: \textit{(1) backgrounds} can dominate the point budget, \textit{(2) small objects} may receive too few points, and \textit{(3) contact patches or tool tips} can be missed. As shown in Fig.~\ref{fig:datapip}, \datapipe instead \textit{(1) extracts} DINOv2~\citep{oquab2023dinov2} patch features and clusters them into entity-level groups, \textit{(2) propagates} these entity identities throughout the clip, and \textit{(3) allocates} a fixed keypoint budget per entity and selects spatially diverse points on each entity's high-visibility frames (Appendix~\ref{subsec:appendix_keypoint_sampling}). The result is a compact query set focused on action-informative entities; a movement filter further marks static or background-dominated tracks so non-moving points do not overwhelm the interaction signal (Appendix~\ref{app:movement_filter}).

\subsection{3D Trace Construction}
\label{subsec:trace_construction}
\noindent\textbf{Global--local reconstruction.}
After keypoint selection, \datapipe must preserve each query's identity and 3D position across long videos despite \textit{(1) egocentric camera motion}, \textit{(2) objects entering or leaving the scene}, and \textit{(3) memory limits of full-video reconstruction}. The middle stage of Fig.~\ref{fig:datapip} addresses these issues with global--local reconstruction: it \textit{(1) uses sparse anchor frames} to establish a shared global coordinate frame, \textit{(2) reconstructs dense local chunks} and aligns them back to that frame, \textit{(3) tracks sampled keypoints} in the common 3D space, and \textit{(4) propagates tracks} across chunk boundaries using each point's last valid world-space position.

\noindent\textbf{Reference-frame traces.}
We reproject the tracks into a per-chunk reference camera to obtain screen-aligned 3D traces $\mathbf{T}_{\mathrm{ref},n}^{t:t+H}=[x_{n,i},y_{n,i},z_{n,i}]_{i=t}^{t+H}$ for each query $n$. This representation \textit{(1) removes camera motion} and \textit{(2) retains image alignment} for the visual backbone (Sec.~\ref{subsec:backbone}). We further normalize trace speed by arc-length reparameterization, reducing duration differences between human and robot demonstrations. Details are provided in Appendix~\ref{subsec:hybrid_vggt} and Appendix~\ref{subsec:progressive_tracking}.
\subsection{Event-Centric Captioning}
\label{sec:caption}
\noindent\textbf{Motion-centric chunking.}
Long demonstrations need language at multiple resolutions: episode captions miss local subgoals, while frame-level captions are expensive and noisy. \datapipe therefore uses traces to define captioning units. In the final stage of Fig.~\ref{fig:datapip}, we smooth per-frame trace acceleration $a_t$ into $\tilde{a}_t$ with a Savitzky--Golay filter~\citep{savitzky1964smoothing}, identify action anchors as prominent peaks $p_i$, and place chunk boundaries at the lowest-acceleration valleys, $b_i = \arg\min_{t \in [p_i, p_{i+1}]} \tilde{a}_t$. This creates short motion-centric events that \textit{(1) limit VLM context length} and \textit{(2) align chunks} with subgoals such as reaching, grasping, moving, and releasing.

\noindent\textbf{Hierarchical VLM captioning.}
For each chunk, the VLM produces captions from \textit{(1) the start frame}, \textit{(2) the midpoint frame}, and \textit{(3) the end frame}, optionally conditioned on a motion mask and an episode-level task description when available. A text-only LLM then merges adjacent captions over sliding windows, yielding both fine-grained captions and coarser task summaries (Appendix~\ref{subsec:appendix_captioning}).
\subsection{Trace Supervision Interface}
\label{subsec:trace_supervision}
Combining semantic queries, reference-frame traces, and event captions, \datapipe converts each video into tuples
    $\mathcal{D}_{\mathrm{TE}}
    = \left\{\left(I_t, l_c, \mathbf{Q}_t, \mathbf{T}_{\mathrm{ref}}^{t-h:t+H}\right)\right\},$
where $I_t$ is the observation, $l_c$ is the event or merged task caption, $\mathbf{Q}_t=\{\mathbf{q}_n^t\}_{n=1}^{N}$ is the query-keypoint set selected at first visibility or carried from history, and $\mathbf{T}_{\mathrm{ref}}^{t-h:t+H}$ contains past and future 3D traces in the reference camera.
Then, $\mu_0$ trained on these tuples learns the prediction map $\mu_0: \left(I_t, l_c, \mathbf{Q}_t, \mathbf{T}_{\mathrm{ref}}^{t-h:t}\right) \mapsto \hat{\mathbf{T}}_{\mathrm{ref}}^{t:t+H},$
which predicts the future motion of the interaction-centric query set.

\begin{figure*}[t!]
    \centering
    \includegraphics[width=1\textwidth]{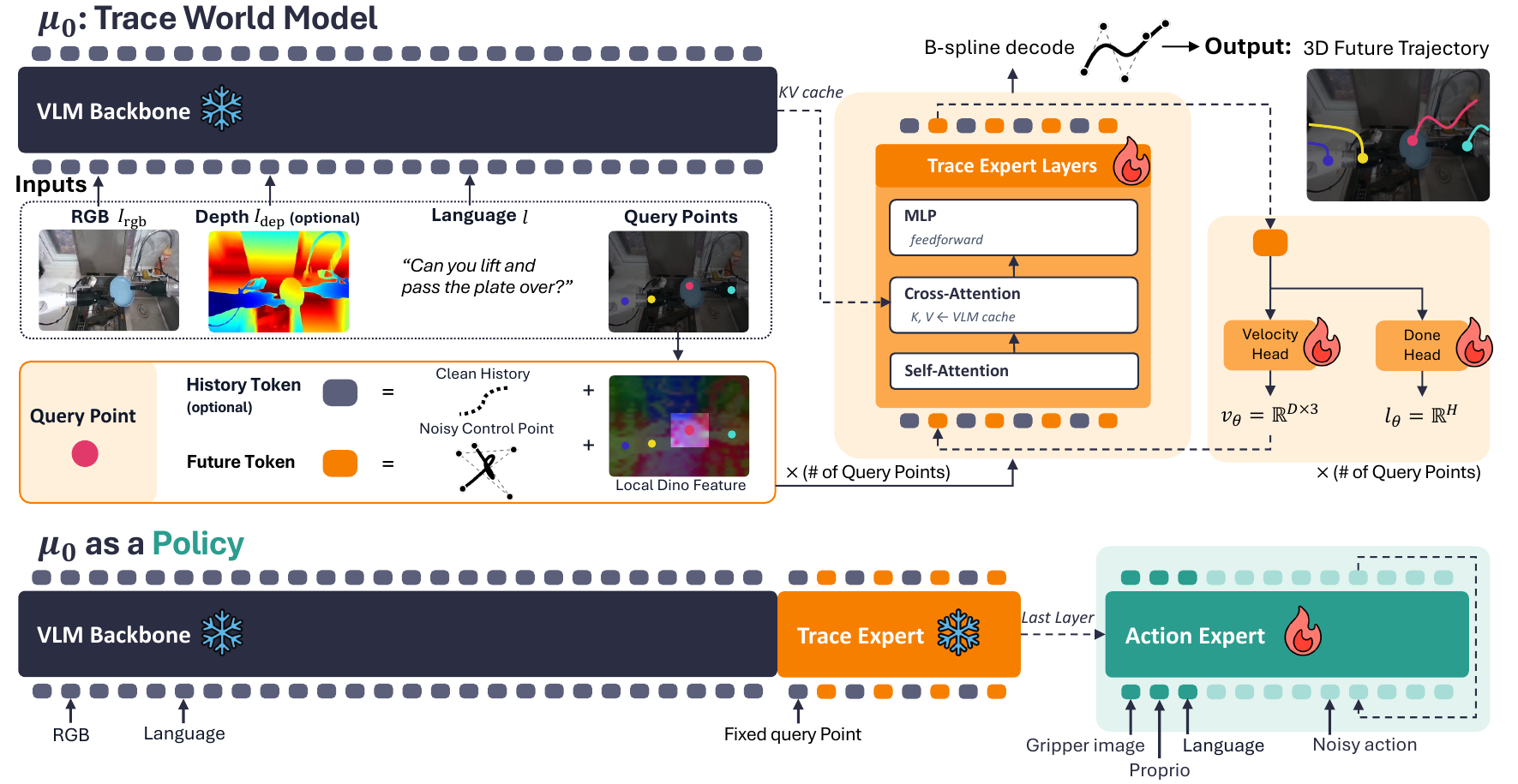}
    \vspace{-1em}
    \caption{\textbf{Overview of \modelname and its action-expert interface.}
    \datapipe provides event-captioned 3D traces for semantic query keypoints. The VLM-conditioned trace context (Sec.~\ref{subsec:backbone}) encodes RGB, language, and optional depth; spline query tokens (Sec.~\ref{subsec:suffix}) represent each keypoint as an exchangeable B-spline query grounded by local DINO features; semantic flow matching (Sec.~\ref{subsec:flow}) denoises control points into smooth future 3D traces; and the action expert (Sec.~\ref{subsec:action_expert}) maps frozen trace features to executable robot actions.
    }
    \vspace{-1em}
    \label{fig:training}
\end{figure*}

\section{\modelname: Query-Conditioned Trace World Model}
\label{sec:training}

\noindent\textbf{Overview.} Using the tuples produced by \datapipe, \modelname learns a query-conditioned dynamics model rather than a pixel generator. It predicts how interaction-centric keypoints move in 3D from \textit{(1) observation}, \textit{(2) language instruction}, and \textit{(3) optional keypoint history}.
This formulation must resolve three coupled challenges: 
\textit{(1) semantic--metric fusion}, retaining large vision-language priors while adding metric 3D reasoning; 
\textit{(2) query equivariance}, handling variable and unordered trace-query sets; and 
\textit{(3) multi-modal dynamics}, representing plausible futures without averaging away contact-rich motion. 
We address these challenges with the three components in Fig.~\ref{fig:training}: 
(\ref{subsec:backbone}) a VLM-conditioned backbone for scene and language context, 
(\ref{subsec:suffix}) a permutation-equivariant trace expert for query-wise spline prediction, and 
(\ref{subsec:flow}) a semantic flow objective for structured future generation. 
Together, these choices turn flexible semantic keypoints into compact metric motion tokens rather than fixed grids or dense scene fields. Downstream action experts can then consume \modelname's trace representation, allowing video-only world-model pretraining to support robot control.

\subsection{Multi-Modal Conditioning Backbone}
\label{subsec:backbone}
Trace prediction requires both global intent and metric scene context: \textit{(1) language} specifies the desired outcome, \textit{(2) RGB} identifies objects and affordances, and \textit{(3) depth} (optional) disambiguates 3D geometry when available.

\noindent\textbf{Semantic reuse.}
As illustrated on the left of Fig.~\ref{fig:training}, we use a pretrained SmolVLM2-2.2B prefix to encode the RGB observation and instruction, then attach a trace expert that cross-attends to the VLM key-value cache while maintaining a separate motion-specific stream~\citep{shukor2025smolvla}. This separates \textit{semantic memory}, preserved by the VLM, from \textit{motion computation}, learned by the trace expert.

\noindent\textbf{Depth pathway.}
Because metric depth is outside the native VLM input space, we route it through \textit{(1) a separate trainable patch stem} before \textit{(2) sharing deeper SigLIP layers} with RGB tokens. This lets the model exploit geometric cues without disrupting pretrained RGB statistics. Architectural and optimization details are in Appendix~\ref{subsec:appendix_backbone}.

\subsection{Permutation-Equivariant Trace Expert}
\label{subsec:suffix}
\noindent\textbf{Exchangeable queries.}
A trace world model should accept arbitrary query keypoints, and its predictions should not depend on the order in which those keypoints are listed. \modelname therefore treats each keypoint as an exchangeable query, matching the query-token block in Fig.~\ref{fig:training}. All queries share the same processing stack, preserving permutation equivariance across the keypoint dimension.

\noindent\textbf{Spline targets.}
For each query, we subtract the current 3D anchor and represent the future as cubic B-spline control points following \citet{liu2025trace}. This target provides \textit{(1) compactness}, replacing dense waypoints with a small control set; \textit{(2) smoothness}, suppressing tracker jitter and high-frequency artifacts; and \textit{(3) easier denoising}, reducing the output dimension.

\noindent\textbf{Query tokenization.}
We tokenize each keypoint's history and noisy future controls as per-query tokens. Each token combines \textit{(1) segment embeddings} for history versus future, \textit{(2) Fourier embeddings} for current pixel location, and \textit{(3) DINO features} for local semantics. Together, these choices ground each token in its visual entity while keeping the query set exchangeable. The VLM backbone and trace expert together form the pretrained \modelname, which downstream action experts reuse as a frozen motion prior. Full target-fitting, tokenization, and DINO-fusion details are in Appendix~\ref{subsec:appendix_trace_tokens}.

\subsection{Flow Matching with Semantic Structure}
\label{subsec:flow}
Even with smooth spline targets, future object motion is uncertain: \textit{(1) multiple paths} can satisfy the same instruction, and \textit{(2) traces} may be truncated or partially occluded. A deterministic regressor would tend to average these futures, yielding traces that are not necessarily actionable.

\noindent\textbf{Conditional denoising.}
We instead train the trace expert as a conditional flow model over B-spline control points, shown in the denoising block of Fig.~\ref{fig:training}. Starting from noisy control points, the model predicts the velocity field toward clean controls under \textit{(1) VLM context}, \textit{(2) per-query token conditions}, and \textit{(3) flow-time modulation} injected with adaLN-Zero~\citep{peebles2023scalable}.

\noindent\textbf{Structural constraints.}
The objective adds two terms for controllable traces: \textit{(1) validity prediction}, which identifies when a keypoint trajectory should terminate under occlusion or track loss, and \textit{(2) semantic rigidity}, which encourages keypoints within the same DINO cluster to preserve local geometry. The training loss is
$
    \mathcal{L}
    = \mathcal{L}_{\text{flow}}
    + \lambda_{\text{done}}\mathcal{L}_{\text{done}}
    + \lambda_{\text{rig}}\mathcal{L}_{\text{rig}},
$
where $\mathcal{L}_{\text{flow}}$ matches the control-point velocity field, $\mathcal{L}_{\text{done}}$ supervises per-step trajectory validity, and $\mathcal{L}_{\text{rig}}$ preserves local geometry within DINO clusters. At inference, $\mu_0$ runs a denoising loop to decode the control points and reconstruct smooth 3D traces from them. The full objective and inference equations are in Appendix~\ref{subsec:appendix_flow_objective}.
\subsection{Trace-Conditioned Action Expert}
\label{subsec:action_expert}
\noindent\textbf{Embodiment transfer.}
\modelname is pretrained from \datapipe video supervision, but robot execution requires actions in a target embodiment. As shown on the right of Fig.~\ref{fig:training}, we freeze the pretrained \modelname---comprising \textit{(1) the VLM backbone} and \textit{(2) the trace expert}---then train only an action expert. This makes the pretrained \modelname reusable across action experts while keeping the learned 3D motion prior embodiment agnostic and limiting action supervision to the target robot interface.

\noindent\textbf{Policy interface.}
The policy uses frozen trace-denoising features as intermediate motion tokens rather than requiring a complete rollout or inverse-kinematics replay at every control step. Specifically, it \textit{(1) reads} features from a single partial-denoising step of  $\mu_0$, \textit{(2) injects} them into VLM features via gated cross-attention, and \textit{(3) predicts} continuous action chunks with an action denoiser conditioned on gripper-camera, proprioception, and language tokens. Details are in Appendix~\ref{subsec:appendix_action_expert}.

\section{Experiment}

\subsection{\modelname Trace Prediction Quality}
\label{subsec:trace_generation}
In this section, we evaluate the trace prediction quality of \modelname both qualitatively and quantitatively, comparing our method against several 2D and 3D baselines.

\noindent\textbf{Evaluation metrics.}
Following \cite{huang2026pointworld}, we compute metrics exclusively on moving points. We evaluate trajectory prediction using Average Displacement Error (ADE) and Final Displacement Error (FDE) \cite{thakkar2026forecasting}. To further evaluate trajectory shape independent of temporal misalignments, we utilize Dynamic Time Warping (DTW). 

\noindent\textbf{Evaluation set.}
We evaluate on 129 slots (16 query keypoints each) from five sources held out from pretraining: Human, Mobile ALOHA, Franka, UR3, and VIMA-Bench. All methods receive the same query keypoints, selected using the training-time movement filter (Appendix~\ref{app:movement_filter}), and metrics use raw ground-truth traces rather than our B-spline-fitted targets (details in Appendix~\ref{sec:appendix_eval_set}).

\begin{table*}[t]
\definecolor{traceTimeShade}{RGB}{242,247,255}
\centering
\caption{\textbf{2D and 3D trace prediction evaluation}. Comparison of trajectory prediction quality over time horizons $T \in \{8, 16, 32\}$. The shaded column reports inference time for trace prediction on one image, with $^{\dagger}$ denoting API latency. All baselines receive the same image and text pairs, except $^{\ddagger}$ which requires depth input.}
\label{tab:ade_fde_dtw_results}
\resizebox{\textwidth}{!}{%
\begin{tabular}{cl ccc ccc ccc ccc ccc ccc >{\columncolor{traceTimeShade}}c}
\toprule
& \multirow{2}{*}{\textbf{Method}} & \multicolumn{3}{c}{\textbf{top1-ADE $\downarrow$}} & \multicolumn{3}{c}{\textbf{top5-ADE $\downarrow$}} & \multicolumn{3}{c}{\textbf{top1-FDE $\downarrow$}} & \multicolumn{3}{c}{\textbf{top5-FDE $\downarrow$}} & \multicolumn{3}{c}{\textbf{top1-DTW $\downarrow$}} & \multicolumn{3}{c}{\textbf{top5-DTW $\downarrow$}} & \multirow{2}{*}{\textbf{Inf. Time $\downarrow$}} \\
\cmidrule(lr){3-5} \cmidrule(lr){6-8} \cmidrule(lr){9-11} \cmidrule(lr){12-14} \cmidrule(lr){15-17} \cmidrule(lr){18-20}
& & $T=8$ & $16$ & $32$ & $8$ & $16$ & $32$ & $8$ & $16$ & $32$ & $8$ & $16$ & $32$ & $8$ & $16$ & $32$ & $8$ & $16$ & $32$ & \\
\midrule
\multirow{6}{*}{\textbf{2D}} 
& Gemini 3.1 Pro & 0.190 & 0.274 & 0.305 & 0.161 & 0.232 & 0.253 & \textbf{0.311} & 0.425 & 0.424 & 0.254 & 0.321 & 0.311 & \textbf{0.183} & 0.258 & 0.284 & 0.152 & 0.208 & 0.224 & 78s$^{\dagger}$ \\
& Gemini 3 Flash & \textbf{0.187} & \textbf{0.271} & 0.299 & 0.158 & 0.231 & 0.254 & 0.312 & 0.414 & 0.405 & 0.252 & 0.329 & 0.316 & \textbf{0.183} & 0.260 & 0.281 & 0.150 & 0.211 & 0.227 & 62s$^{\dagger}$ \\
& GPT-5.5        & 0.199 & 0.281 & 0.307 & 0.178 & 0.249 & 0.272 & 0.329 & 0.411 & \textbf{0.404} & 0.284 & 0.344 & 0.329 & 0.196 & 0.274 & 0.299 & 0.173 & 0.238 & 0.259 & 38s$^{\dagger}$ \\
& Track2Act \cite{bharadhwaj2024track2act}      & 0.209 & 0.311 & 0.369 & 0.190 & 0.262 & 0.293 & 0.350 & 0.493 & 0.555 & 0.287 & 0.351 & 0.346 & 0.206 & 0.303 & 0.358 & 0.181 & 0.245 & 0.270 & 0.85s \\
& Hamster \cite{li2025hamster}        & 0.202 & 0.276 & \textbf{0.297} & 0.178 & 0.239 & 0.256 & 0.326 & \textbf{0.400} & 0.411 & 0.274 & 0.320 & 0.330 & 0.197 & 0.261 & \textbf{0.277} & 0.170 & 0.220 & 0.233 & 14.4s \\
& \textbf{\modelname (Ours)} & 0.202 & 0.279 & 0.315 & \textbf{0.124} & \textbf{0.188} & \textbf{0.227} & 0.322 & 0.410 & 0.447 & \textbf{0.186} & \textbf{0.261} & \textbf{0.284} & 0.184 & \textbf{0.254} & 0.296 & \textbf{0.114} & \textbf{0.171} & \textbf{0.211} & \textbf{0.29s} \\
\midrule
\multirow{5}{*}{\textbf{3D}} 
& 3DFlowAction \cite{zhi20253dflowaction}   & 0.615 & 0.692 & 0.716 & 0.531 & 0.605 & 0.630 & 0.753 & 0.819 & 0.818 & 0.648 & 0.714 & 0.712 & 0.614 & 0.688 & 0.711 & 0.529 & 0.600 & 0.623 & 3.38s \\
& Dream2Flow$^{\ddagger}$  \cite{dharmarajan2026dream2flow}    & 0.354 & 0.451 & 0.505 & 0.201 & 0.286 & 0.336 & 0.497 & 0.616 & 0.660 & 0.287 & 0.378 & 0.403 & 0.352 & 0.449 & 0.500 & 0.198 & 0.281 & 0.329 & 106.8s \\
& TraceGen$^{\ddagger}$ \cite{lee2026tracegen}     & 0.327 & 0.416 & 0.464 & 0.208 & 0.276 & 0.325 & 0.478 & 0.548 & 0.642 & 0.267 & 0.329 & 0.370 & 0.298 & 0.375 & 0.413 & 0.204 & 0.262 & 0.299 & 1.20s \\
& \textbf{\modelname(Ours)}  & \textbf{0.209} & \textbf{0.288} & \textbf{0.325} & \textbf{0.132} & \textbf{0.199} & \textbf{0.239} & \textbf{0.331} & \textbf{0.425} & \textbf{0.464} & \textbf{0.200} & \textbf{0.278} & \textbf{0.305} & \textbf{0.191} & \textbf{0.263} & \textbf{0.308} & \textbf{0.127} & \textbf{0.187} & \textbf{0.223} & \textbf{0.29s} \\ 
\bottomrule
\end{tabular}%
}
\end{table*}

\begin{figure*}[t!]
    \centering
    \includegraphics[width=\textwidth]{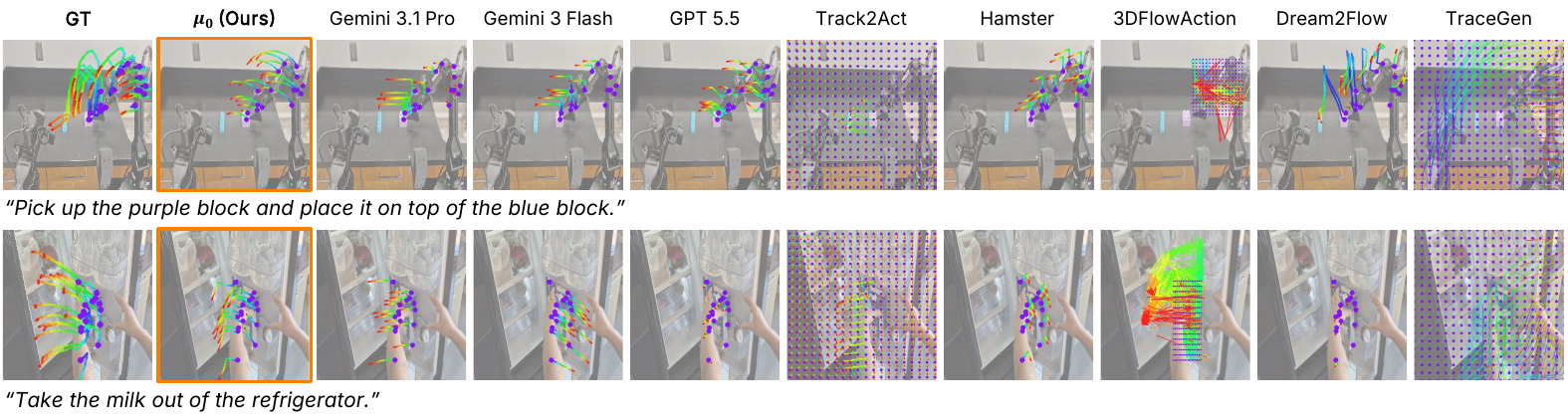}
    \vspace{-1.3em}
    \caption{\textbf{Qualitative comparison of predicted traces.} We compare predicted traces from \modelname{} and baselines on two manipulation tasks. \modelname{} produces coherent and goal-directed traces, while avoiding the noisy or misaligned predictions observed in prior methods (more results in  Appendix~\ref{sec:appendix_trace_qual}).} 
    \label{fig:qual}
\end{figure*}

\noindent\textbf{Results.}
Table~\ref{tab:ade_fde_dtw_results} shows that \modelname consistently improves multi-sample trace prediction quality. \textbf{(1)} In 2D, \modelname achieves the best Top-5 ADE, FDE, and DTW across all horizons. These gains indicate that its sampled traces contain more accurate goal-directed futures even when Top-1 performance is competitive with strong VLM baselines. \textbf{(2)} In 3D, \modelname obtains the best result on every reported ADE, FDE, and DTW metric across all horizons. \textbf{(3)} Beyond accuracy, \modelname is also efficient: its 0.29s prediction latency is 2.9$\times$ faster than the next-fastest reported 2D baseline (Track2Act \cite{bharadhwaj2024track2act}, 0.85s). \textbf{(4)} The qualitative examples in Figure~\ref{fig:qual} further support these trends. \textbf{(5)} Corrupting 41\% of caption words or dropping/jittering 30\% of input keypoints degrades these metrics by 2.9--3.3\% and 1.2--4.1\%, respectively (Appendix~\ref{sec:appendix_robustness}).

\subsection{Action Generation with Pretrained \modelname under Both Simulated and Real-World Scenarios}
\label{subsec:action_generation}

In this section, we evaluate whether pretrained \modelname{} can serve as a motion prior for action generation.

\noindent\textbf{Simulated experiment setup.}
We evaluate each method on 8 representative tasks in RoboCasa365~\citep{nasiriany2026robocasa365}, a large-scale simulation benchmark for everyday kitchen manipulation that randomizes scene layouts, object instances, and initial configurations (details are in Appendix~\ref{sec:appendix_robocasa}). We benchmark our method against three classes of baselines: \textbf{(1) Diffusion Policy}~\citep{chi2025diffusion}, trained from scratch on target-domain demonstrations; \textbf{(2) action-labeled VLAs}, $\pi_0$~\citep{black2025pi0} and $\pi_{0.5}$~\citep{intelligence2025pi_}, pretrained with large-scale robot action labels; and \textbf{(3) video-only trace models}, TraceGen~\citep{lee2026tracegen}, pretrained without proprioceptive or action supervision, like $\mu_0$. For all pretrained methods, we fully finetune only the action expert on the RoboCasa365 data.

\begin{table*}[t!]
\centering
\caption{\textbf{Simulation results in RoboCasa365.} We evaluate downstream action generation on 8 representative RoboCasa365 tasks and report success rates (\%). \textbf{Bold} and \underline{underline} numbers indicate the best and second-best results in each row, respectively.}
\label{tab:robocasa_results}
\resizebox{\textwidth}{!}{%
\begin{tabular}{l c cc cc}
\toprule
\multirow{2}{*}{\textbf{Task}} & \textbf{No pretraining} & \multicolumn{2}{c}{\textbf{Action-labeled pretraining (VLA)}} & \multicolumn{2}{c}{\textbf{Video-only pretraining}} \\
\cmidrule(lr){2-2} \cmidrule(lr){3-4} \cmidrule(lr){5-6}
& Diffusion Policy~\citep{chi2025diffusion} & $\pi_0$~\citep{black2025pi0} &  $\pi_{0.5}$~\citep{intelligence2025pi_} & TraceGen~\citep{lee2026tracegen} + action expert & Ours (\modelname + action expert) \\
\midrule
CloseFridge                 & 34 & \underline{44} & 34 & 38 & \textbf{54} \\
OpenFridge                  & \underline{28} & 12 & 26 & \textbf{36} & 18 \\
CoffeeServeMug              & 28 & 34 & \textbf{48} & \underline{42} & 36 \\
PickPlaceFridgeShelfToDrawer& 28 & 30 & \textbf{66} & 30 & \underline{40} \\
TurnOnMicrowave             & 0 & 2 & \textbf{12} & 0 & \underline{4}  \\
SlideToasterOvenRack        & 48 & 46 & \textbf{76} & 28 & \underline{56} \\
PickPlaceCounterToCabinet   & 6 & \underline{18} & \textbf{54} & 0 & 12 \\
TurnOnToasterOven           & 10 & 16 & \underline{20} & 10 & \textbf{22}  \\
\midrule
\textbf{Average Success Rate (\%)} & 22.75 & 25.25 & \textbf{42} & 23 & \underline{30.25} \\
\bottomrule
\end{tabular}%
}
\end{table*}

\noindent\textbf{Results of simulated scenarios.}
Table~\ref{tab:robocasa_results} presents the success rates across the 8 selected RoboCasa tasks. \textbf{(1)} Overall, \modelname{} + action expert achieves a 30.25\% average success rate, outperforming $\pi_0$ by 5.0 points, despite relying solely on video-only pretraining.
\textbf{(2)} At the same time, $\pi_{0.5}$ remains stronger on average; however, this comparison is not data-matched: $\pi_0$ and $\pi_{0.5}$ share an identically sized backbone pretrained on large-scale action-labeled data, which is costly and difficult to scale, whereas \modelname is pretrained only on action-free video. $\pi_0$ and $\pi_{0.5}$ are nonetheless fine-tuned from public pretrained checkpoints, and all methods share the same demonstrations, training steps, and batch size (Appendix~\ref{sec:appendix_robocasa}, Table~\ref{tab:robocasa_training_hyperparameters}); \modelname{}+action expert also has fewer inference-time parameters than either baseline (Appendix~\ref{sec:appendix_deployment_cost}).
\textbf{(3)} Compared with the previous video-only trace baseline (TraceGen), \modelname{} improves average success by 7.25 points, which we expect to reflect the benefit of stronger 3D trace prediction.

\begin{figure*}[t]
    \centering
    \includegraphics[width=\textwidth]{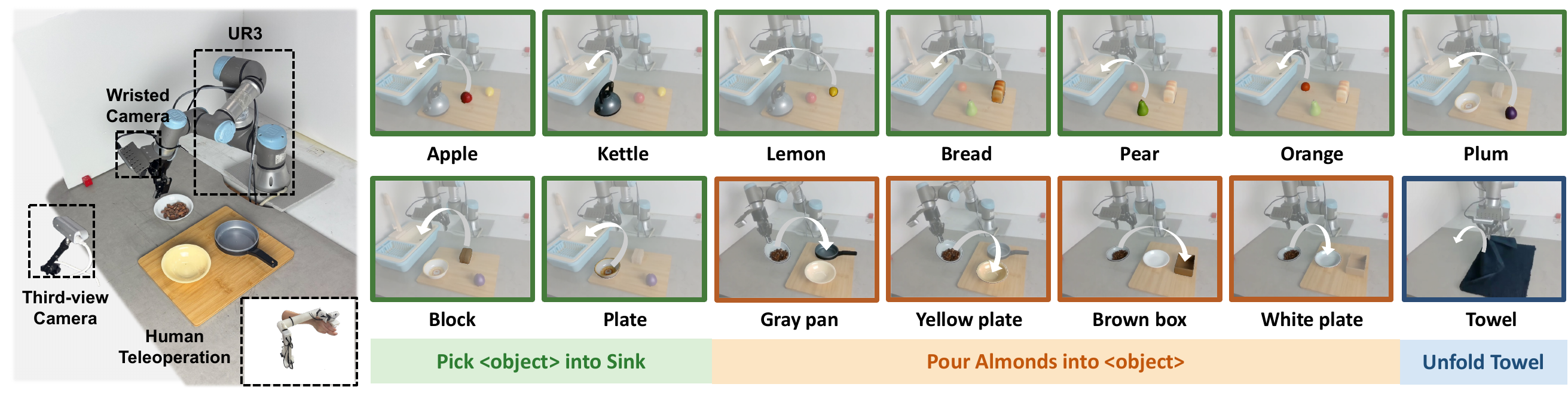}
    \vspace{-2em}
    \caption{\textbf{Real-world experimental setup and task visualizations.} The setup includes a UR3 robot arm with a two-finger gripper and the three real-world manipulation tasks used for evaluation.}
    \label{fig:real_robot_setup}
    \vspace{-1em}
\end{figure*}

\noindent\textbf{Real-world experiment setup.}
Our real-world experiments use a UR3 robot arm equipped with a two-finger gripper, as shown in Figure~\ref{fig:real_robot_setup}. We evaluate each method on three tasks: two multi-instruction tasks, \textit{Pick \textless object\textgreater{} into Sink} and \textit{Pour Almonds into \textless object\textgreater{}}, and a single \textit{Unfold Towel} task. We collect 90, 80, and 50 demonstrations for the three tasks, respectively, and evaluate each task over 20 rollouts. Additional experimental details are reported in Appendix~\ref{sec:appendix_real_robot}.

\begin{figure*}[t]
    \centering
    \includegraphics[width=0.9\textwidth]{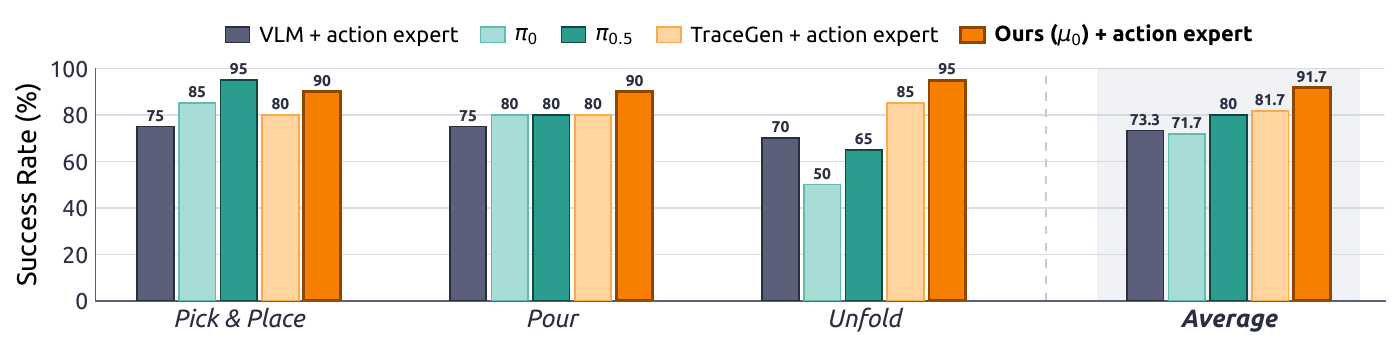}
    \vspace{-1.0em}
    \caption{\textbf{Real-world evaluation results.} Bar charts show average success rates (\%) for three in-distribution UR3 manipulation tasks. Pick \& Place and Pour are averaged over multiple objects.}
    \vspace{-1.0em}
    \label{fig:real_world_results}
\end{figure*}

\noindent\textbf{Results of real-world scenarios.}
\textbf{(1)} Figure~\ref{fig:real_world_results} shows that \modelname{} + action expert achieves the highest average success rate of 91.7\%, outperforming all baselines across the three real-world tasks on average. 
\textbf{(2)} Compared with the VLM + action expert baseline, which keeps the same policy architecture but removes the trace expert, $\mu_0$ shows an 18.4 percentage-point gap, indicating that frozen trace features provide useful motion guidance beyond generic VLM representations. 
\textbf{(3)} \modelname{} + action expert also surpasses the action-labeled VLA baselines $\pi_0$ and $\pi_{0.5}$ by 20.0 and 11.7 percentage points.
\textbf{(4)} Compared with the previous video-only baseline TraceGen, \modelname{} improves average success by 10.0 percentage points, which we attribute to the stronger \datapipe supervision and architecture. \textbf{(5)} \modelname{}+action expert has fewer inference-time parameters than $\pi_0$ or $\pi_{0.5}$, with an end-to-end latency $\sim$16\% higher than $\pi_{0.5}$'s (Appendix~\ref{sec:appendix_deployment_cost}).

\noindent\textbf{Scaling analysis.}
Our scaling results offer two main takeaways. First, trace prediction consistently improves with larger models and more pretraining data, yielding the best top5-DTW with the 2.59B model. Second, the trace representation effectively transfers to robot control: the performance gap between our model and the \textit{w/o Trace} variant widens significantly as the action head size decreases, demonstrating that trace-space pretraining provides crucial motion structure that limited policy capacity cannot recover. Full protocols and results are in Appendix~\ref{subsec:scaling_analysis} (Tables~\ref{tab:scaling} and~\ref{tab:action_head_scaling}).
\noindent\textbf{Design ablations.} 
Component-wise ablations verifying every major choice—including B-spline parameterization, DINOv2 features, rigidity loss, depth input, and historical traces—are detailed in Appendix~\ref{subsec:ablation_trace} (Table~\ref{tab:ablation}).

\section{Related Work}

\topic{World models and visual motion priors.}
Embodiment-agnostic robot learning leverages world models to forecast scene dynamics independently of specific action spaces~\citep{du2023learning, xu2024flowcrossdomainmanipulationinterface, yuan2024general, huang2026pointworld}.
While pixel-space models offer broad visual priors~\citep{wu2024unleashing, guo2026ctrlworld} and world-action models jointly predict frames and actions~\citep{li2026causal, ye2026world, ye2026gigaworld}, they waste capacity on dense appearance rather than geometry and contact.
Intermediate representations like features, 2D tracks, and 3D flow mitigate this~\citep{jang2026lace, zhou2025dinowm, hu2025video, gu2024rt, bharadhwaj2024track2act, wen2024any, vecerik2024robotap, kambara2026lilac, zhi20253dflowaction, wang2026lamp}, yet they carry distinct limitations: latent features lack control, 2D tracks lose metric depth, and fixed grids waste budget on backgrounds.
Instead, \modelname predicts explicit 3D trajectories for query-selected interaction points, providing a compact, metric, and reusable motion interface.

\topic{Trace-guided manipulation.}
Visual motion plans for manipulation generally fall into three categories: 
(i) VLM-based waypoint or end-effector tracking~\citep{li2025hamster, zhou2025robotracer, yuan2024robopoint, yang2025magma}, 
(ii) video generation followed by track extraction~\citep{ko2024learning, bharadhwaj2025genact, li2025novaflow, dharmarajan2026dream2flow}, and 
(iii) policies directly predicting tracks via diffusion or flow matching~\citep{nguyen2026pixel, gao2025flip, lin2026roboflow4d}.
TraceGen~\citep{lee2026tracegen} is closest to our work but relies on fixed-grid traces and requires inference-time depth. Extended discussions are in Appendix \ref{sec:appendix_related_work}.

\section{Conclusion}
\label{sec:conclusion}

We introduced \modelname, a query-conditioned 3D trace-space world model for cross-embodiment manipulation. Instead of predicting pixels or embodiment-specific actions, \modelname predicts smooth future 3D motion for semantically selected interaction keypoints. Its supervision comes from \datapipe, which turns heterogeneous videos into event-captioned 3D trace tuples through semantic keypoint selection, globally aligned tracking, and motion-centric captioning. After video-only pretraining, the frozen trace model can be reused by action experts, providing an embodiment-agnostic motion prior for downstream robot control. Across trace forecasting, simulation, and real-world robot experiments, our results support 3D interaction traces as a compact and actionable representation for scalable robot world modeling.

\textbf{Limitations and Future Work.}
\modelname inherits errors from the perception stack used to construct traces: failures in semantic clustering, 3D reconstruction, tracking, or captioning can produce noisy supervision.
The trace representation captures geometry and motion but does not explicitly model forces, tactile feedback, or contact modes, which may be important for fine manipulation. Our action expert evaluations focus on tabletop manipulation with limited embodiments and task families; broader validation on mobile manipulators, dexterous hands, and longer-horizon tasks remains future work.

\clearpage
\acknowledgments{Lee, Jung and Huang are supported by DARPA HR001124S0029-AIQ-FP-019, National Science Foundation TRAILS Institute (2229885). Private support was provided by Open Philanthropy and Apple. The authors acknowledge the National Artificial Intelligence Research Resource (NAIRR) Pilot for contributing to this research result. This work was supported by Samsung Research Funding \& Incubation Center of Samsung Electronics under Project Number SRFC-IT2402-17. We thank Jonguk Cheon and Seokjin Park for their help and support with this project.}

\bibliography{example}  %

\newpage
\appendix
\let\addcontentsline\oldaddcontentsline
\renewcommand{\contentsname}{\Large{Appendix}}
\setcounter{tocdepth}{4}
\tableofcontents
\clearpage
\section{Dataset Construction}

\subsection{Semantic Keypoint Sampling Details}
\label{subsec:appendix_keypoint_sampling}

For each chunk, we compute DINOv2~\cite{oquab2023dinov2} patch descriptors on a small set of representative frames and cluster the descriptors into entity-level groups. Cluster identities are propagated temporally by bipartite matching between adjacent frames, where the matching score combines feature similarity and spatial overlap. Given a per-chunk budget of $N$ keypoints, each entity receives a quota proportional to its visible patch coverage, with a minimum allocation for small entities that remain salient but occupy few patches. Final keypoints are selected by farthest-point sampling within each entity mask on frames where the entity is visible, producing spatially diverse query points that are less likely to fall on background or transient occluders. The resulting DINO cluster identity is stored with each keypoint and reused as the part label for the rigidity loss in Appendix~\ref{subsec:appendix_flow_objective}.

\subsection{Movement Filter}
\label{app:movement_filter}

A substantial fraction of tracked points correspond to background or static structures that carry no task-relevant information and bias the model toward predicting zero motion. For each keypoint $i$, we project its trace through $\mathrm{cam}_{\mathrm{ref}}$ to obtain $(u_i^t, v_i^t, z_i^t)$, weight depth by $\lambda_z = 0.1$ so that pixel motion dominates, and compute the trace diameter
$d_i = \max_{t,t' \in \mathcal{V}_i}\|(u_i^t{-}u_i^{t'},\,v_i^t{-}v_i^{t'},\,\lambda_z(z_i^t{-}z_i^{t'}))\|_2$
over the visible frame set $\mathcal{V}_i$. A keypoint is marked \emph{moving} if $d_i$ exceeds a resolution-adaptive threshold $\tau_m = \sqrt{(H/20)^2 + (W/20)^2}$, where $H$ and $W$ denote the frame height and width in pixels. We use the maximum pairwise displacement over the trajectory, rather than instantaneous velocity, as it captures the full extent of motion while remaining robust to per-frame tracker jitter.

\subsection{Hybrid Global--Local 3D Reconstruction}
\label{subsec:hybrid_vggt}

Globally consistent depth, intrinsics, and extrinsics are the prerequisite for placing every 3D trace in a single reference camera frame. \datapipe uses a hybrid VGGT~\cite{vggt} scheme that combines one global sparse pass with dense local passes, enabling long-horizon manipulation videos to be processed without fitting the entire sequence in memory.

\noindent\textbf{Global sparse pass.}
Given a video of length $T_{\text{total}}$, we uniformly subsample at most $T_{\text{sparse}}$ anchor frames and feed them through VGGT in a single forward call, yielding extrinsics $\{\mathbf{E}_t^{\text{sparse}}\}_{t \in \mathcal{S}}$ in a \emph{global frame} together with a $\mathbf{K}^{\text{global}}$ obtained by averaging the per-frame intrinsics. Per-chunk intrinsics introduce visible discontinuities at chunk boundaries, so a single shared $\mathbf{K}^{\text{global}}$ is essential.

\noindent\textbf{Dense passes and SE(3) alignment.}
The full video is split into non-overlapping chunks of $T_{\text{chunk}}$ frames, each producing chunk-local depth $\mathbf{D}^{(c)}$ and extrinsics $\{\mathbf{E}_t^{(c)}\}$. For every chunk $c$, the anchor frames in $\mathcal{S}\cap c$ act as shared observations, and we solve for the rigid transform $\mathbf{A}^{(c)} \in \mathrm{SE}(3)$ that maps chunk-local poses to global poses,
\begin{equation}
    \mathbf{A}^{(c)} = \arg\min_{\mathbf{A} \in \mathrm{SE}(3)} \sum_{t \in \mathcal{S}\cap c} \big\| \mathbf{A}\,\mathbf{E}_t^{(c)} - \mathbf{E}_t^{\text{sparse}} \big\|^2.
\end{equation}
Because each chunk aligns \emph{directly} to the same global anchors rather than to its predecessor, alignment errors are independent and bounded across chunks instead of compounding.

\subsection{Progressive 3D Tracking Across Chunks}
\label{subsec:progressive_tracking}

Running a 3D point tracker independently per chunk discards continuity: the same physical point would be re-discovered with a new identity in every chunk, and any object missed by DINO in one chunk would simply vanish. We instead track \emph{progressively}. The first chunk is processed by feeding peak-frame keypoints through TAPIP3D~\cite{zhang2025tapip3dtrackingpointpersistent}, which produces 3D world-space coordinates $\{\mathbf{p}_i^t\}$ and visibility flags. For chunk $c \geq 1$, every active group is propagated by using its last known 3D world position in previous chunk as a 3D query at the first frame of chunk $c$; because positions live in the same global frame, propagation operates on world-space 3D coordinates and is therefore robust to the large camera motion typical of egocentric video.

\subsection{Event-Centric Captioning Details}
\label{subsec:appendix_captioning}

Given tracked traces, we compute a scalar motion signal by averaging per-frame accelerations over valid moving keypoints. The signal is smoothed into $\tilde{a}_t$ with a Savitzky--Golay filter~\cite{savitzky1964smoothing}, and prominent local maxima $p_i$ are treated as action anchors. Chunk boundaries are placed at low-motion transition points,
\begin{equation}
    b_i = \arg\min_{t \in [p_i, p_{i+1}]} \tilde{a}_t,
\end{equation}
with minimum- and maximum-duration constraints to avoid degenerate clips. For each chunk, the VLM receives the start, midpoint, and end frames, together with an optional motion mask rendered from the moving traces and an optional episode-level task description. It produces a structured caption describing the object state at the beginning, the interaction that occurs, and the state change at the end. A text-only LLM then merges adjacent chunk captions over sliding windows, yielding paired frame ranges for both fine-grained event captions and coarser task summaries.

\subsection{\datapipe Runtime and Compute Cost}
\label{subsec:appendix_runtime}

We measure the end-to-end throughput of the full \datapipe pipeline---keypoint selection, global--local 3D reconstruction, progressive 3D tracking, and hierarchical captioning combined---on representative manipulation videos. \datapipe processes video at approximately 2.4--3.3 frames/s on a single GPU, with 3D tracking (Appendix~\ref{subsec:progressive_tracking}) accounting for 45--58\% of total runtime, the largest single cost. Runtime scales near-linearly with the number of processed frames, since chunk-wise reconstruction and tracking avoid the memory and compute blowup of processing an entire video at once. At this throughput, extracting trace supervision for our full $\sim$13M-frame training corpus requires approximately 1.2K single-GPU hours; because each video is processed independently, this cost is fully parallelizable across a GPU cluster rather than a serial bottleneck.

\subsection{Pretraining Data Composition}
\label{subsec:appendix_data_composition}

Table~\ref{tab:pretraining_data} summarizes the four video sources processed by \datapipe to construct the \modelname pretraining corpus. Approximately 52\% of the processed frames come from human videos (EgoDex, EgoVerse) and 48\% from robot videos (Agibot, DROID). At the point-tracking sampling rates of 5 fps for human videos and 10 fps for robot videos, this corresponds to roughly 16 days of human video and 7 days of robot video, for a total of approximately 23 days of wall-clock video.

\begin{table*}[h]
\centering
\caption{\textbf{Pretraining data composition.} Source datasets processed by \datapipe to construct the \modelname pretraining corpus.}
\label{tab:pretraining_data}
\resizebox{0.75\textwidth}{!}{%
\begin{tabular}{lccccc}
\toprule
\textbf{Dataset} & \textbf{Modality} & \textbf{Episodes} & \textbf{Frames} & \textbf{Storage} & \textbf{License} \\
\midrule
EgoDex~\cite{hoque2026egodex} & Human & 51.8K & 4.1M & 4 TB & CC-BY-NC-ND-4.0 \\
EgoVerse~\cite{punamiya2026egoverse} & Human & 3.4K & 2.7M & 2.7 TB & CC BY-SA 4.0 \\
Agibot~\cite{bu2025agibot_iros} & Robot & 123K & 3.7M & 8 TB & CC BY-NC-SA 4.0 \\
DROID~\cite{khazatsky2024droid} & Robot & 25.4K & 2.5M & 1 TB & CC-BY-4.0 \\
\midrule
\textbf{Total} & & \textbf{203.6K} & \textbf{13.0M} & \textbf{15.7 TB} & \\
\bottomrule
\end{tabular}%
}
\end{table*}

\section{Architecture and Training Details}

\subsection{Backbone and Optimization Details}
\label{subsec:appendix_backbone}

The conditioning backbone begins with a pretrained SmolVLM2-2.2B model acting as a vision-language prefix, truncated to its first $L_{\text{vlm}} = 20$ text-decoder layers. The Trace Expert has the same depth ($20$ layers) with hidden width scaled to $0.5\times$ that of the VLM. Following \citet{shukor2025smolvla}, the Trace Expert interleaves cross-attention against the VLM key-value cache with self-attention every two layers. The inputs to the VLM prefix are an RGB image $I_{\text{rgb}}$, an optional metric depth map rendered as an RGB image $I_{\text{dep}}$, and a tokenized textual instruction $l$; both image modalities are resized to $512\times512$.

To incorporate metric depth without disrupting pretrained RGB visual statistics, $I_{\text{dep}}$ is normalized through a Turbo colormap and routed through a separate trainable patch-embedding stem cloned from the RGB stem at initialization. RGB and depth tokens then share the subsequent deeper SigLIP layers, allowing the network to adapt to depth statistics while preserving a unified visual representation. RGB frames pass through ColorJitter with strength $s{=}0.3$; depth is augmented in the meter domain with zero-mean Gaussian noise of standard deviation $\sigma_{d}{=}0.01$\,m \emph{before} the Turbo colormap, preserving the meter-to-color mapping. We optimize with AdamW at base learning rate $10^{-4}$ and weight decay $10^{-10}$, gradient clipped to norm $10$. Training runs for $2{\times}10^{5}$ steps with gradient checkpointing, an effective batch size of $24$ across two GPUs ($6$ per GPU), and $N$ uniformly sampled from $[1, 256]$ keypoints per sample. The VLM and SigLIP tower for RGB is frozen from SmolVLM2-2.2B; the trace expert, trace projections, embedding tables, the depth-only stem, and the adaLN-Zero heads are randomly initialized, with the adaLN-Zero output Linears and the uv-MLP output Linear zero-initialized so the model begins at a well-conditioned step-zero identity.

\subsection{Trace Target and Tokenization Details}
\label{subsec:appendix_trace_tokens}

The Trace Expert consumes three slices of the \datapipe trace $\mathbf{T}_{\mathrm{ref}}^{t-h:t+H}$ in $\mathrm{cam}_{\mathrm{ref}}$: a past history $\mathbf{H} \in \mathbb{R}^{N\times h\times 3}$, a current anchor $\mathbf{c} \in \mathbb{R}^{N\times 3}$ at frame $t$, and a future target $\mathbf{T}^{1} \in \mathbb{R}^{N\times H\times 3}$, with $h{=}8$ and $H{=}32$. We subtract the anchor from history ($\mathbf{H} \leftarrow \mathbf{H}-\mathbf{c}$) and predict an anchor-relative, per-axis-rescaled future
\begin{equation}
    \tilde{\mathbf{T}}^{1}_{n,k}
    = (\mathbf{T}^{1}_{n,k}-\mathbf{c}_{n}) / \boldsymbol{s}_{\Delta},
    \label{eq:delta_target_appendix}
\end{equation}
where $\boldsymbol{s}_{\Delta}$ is a per-axis 95th-percentile scale precomputed once over the training corpus. Anchor-relative targets remove the slow scene-coordinate component and match the variance to the unit-Gaussian noise prior.

Rather than regress the $H$-step anchor-relative future directly, we re-parameterize each keypoint's future as a degree-3 B-spline with $D{=}10$ control points. The anchor-prepended scaled future $[\mathbf{0};\tilde{\mathbf{T}}^{1}_{n}]\in\mathbb{R}^{(H+1)\times3}$ is fit in the dataloader by row-weighted ridge least squares,
\begin{equation}
    \mathbf{P}^{\star}_{n}
    = \arg\min_{\mathbf{P}\in\mathbb{R}^{D\times 3}}
    \big\| \mathbf{M}_{n} \odot (\mathbf{B}\mathbf{P} - [\mathbf{0};\tilde{\mathbf{T}}^{1}_{n}]) \big\|_{F}^{2}
    + \lambda_{\text{bsp}}^{2}\big\|\boldsymbol{\Gamma}\mathbf{P}\big\|_{F}^{2},
    \label{eq:bspline_fit_appendix}
\end{equation}
where $\mathbf{B}\in\mathbb{R}^{(H+1)\times D}$ is a fixed cubic B-spline basis sampled on a uniform grid with the curve pinned at the anchor ($t{=}0$), the per-step row weight $\mathbf{M}_{n}$ zeros invalid future steps so they exert no pull on $\mathbf{P}$, and $\boldsymbol{\Gamma}$ is the first-order finite-difference operator on consecutive control points. The Tikhonov term with $\lambda_{\text{bsp}}{=}0.2$ gently equalizes control-point spacing and compresses the flow-matching target distribution, and an element-wise post-fit clip $|\mathbf{P}^{\star}|\leq1.5$ bounds the target box. A keypoint participates in the flow loss only when at least $D$ of its $H$ future steps are valid; otherwise it is dropped from the flow loss entirely. The network's flow-matching target is $\mathbf{P}^{\star}\in\mathbb{R}^{N\times D\times3}$, and rollouts decode in a single matrix multiply $\hat{\mathbf{T}}^{1}=\mathbf{B}\hat{\mathbf{P}}$ with the anchor row stripped.

We adopt a flat tokenization that splits the trace stream into a clean-history segment and a noisy control-point segment, each with its own grouping axis. History is grouped along time at $g_{\text{time}}{=}h$, yielding $G_{\text{hist}}{=}h/g_{\text{time}}{=}1$ token per keypoint; control points are grouped along the control-point axis at $g_{\text{cp}}{=}D$, yielding $G_{\text{cp}}{=}D/g_{\text{cp}}{=}1$ token per keypoint. Thus, each keypoint contributes $G=G_{\text{hist}}+G_{\text{cp}}$ tokens indexed by $j$. Two separate linear lifts $W_{\text{hist}}: \mathbb{R}^{3g_{\text{time}}}\to\mathbb{R}^{D_{\text{exp}}}$ and $W_{\text{cp}}: \mathbb{R}^{3g_{\text{cp}}}\to\mathbb{R}^{D_{\text{exp}}}$ produce the base token $\mathbf{e}_{n,j}$. We add three positional components: a learned group-index embedding, a binary segment embedding separating history from future, and a 2D Fourier expansion of the current-frame $(u,v)$ passed through a zero-initialized uv-MLP. RoPE positions for all suffix tokens are pinned to the prefix-end index, so rotary attention contributes no positional signal between suffix tokens; all temporal and spatial information lives in the additive embeddings. Within the suffix, attention is causal across the history sub-block and bidirectional elsewhere---control-point tokens attend to each other and to all history---which keeps the keypoint axis exchangeable.

While the VLM encodes global scene-level context, predicting precise traces requires sharp, location-specific cues for each query point. Inspired by \citet{thakkar2026forecasting}, we sample a frozen DINO-base feature map at each keypoint's current-frame pixel coordinate via bilinear grid sampling. This yields a localized semantic feature $\mathbf{f}_{n}^{\text{dino}}\in\mathbb{R}^{D_{\text{dino}}}$ for keypoint $n$. A two-layer MLP fuses this descriptor into each trace token associated with the keypoint,
\begin{equation}
    \mathbf{e}_{n,j}
    \leftarrow W_2\operatorname{SiLU}\!\left(W_1\operatorname{concat}(\mathbf{e}_{n,j},\mathbf{f}_{n}^{\text{dino}})\right),
    \label{eq:dino_fuse_appendix}
\end{equation}
where $W_1$ and $W_2$ are learnable weights and $j\in\{\text{hist},\text{cp}\}$. This directly injects part-level semantic priors at the token level, bridging global scene context with localized point statistics.

\subsection{Flow-Matching Objective and Inference}
\label{subsec:appendix_flow_objective}

We train the Trace Expert to generate target B-spline control points $\mathbf{P}^{\star}$ using conditional flow matching. Given standard Gaussian noise $\boldsymbol{\epsilon}\sim\mathcal{N}(\mathbf{0},\mathbf{I})$ and virtual time $\tau\in[0,1]$, the linear probability path is
\begin{equation}
    \mathbf{P}^{\tau}=\tau\boldsymbol{\epsilon}+(1-\tau)\mathbf{P}^{\star}.
\end{equation}
The network $v_{\theta}$ predicts the constant-in-time target velocity $\boldsymbol{\epsilon}-\mathbf{P}^{\star}$ that transports noise to clean data. To condition the architecture on the flow time step, we route $\tau$ through an adaLN-Zero module at each Trace Expert layer. A shared sinusoidal embedding encodes $\tau$ and emits layer-specific shift, scale, and gate vectors for both the attention and MLP sublayers. The final linear layer of each conditioning head is initialized to zero, so each expert block acts as an exact identity function at initialization.

The primary flow loss is the masked mean squared error of the predicted velocity in control-point space,
\begin{equation}
    \mathcal{L}_{\text{flow}}
    = \mathbb{E}_{\tau,\boldsymbol{\epsilon}}
    \left[
    \left\|v_{\theta}(\mathbf{P}^{\tau},\tau,F_{\text{cond}})
    - (\boldsymbol{\epsilon}-\mathbf{P}^{\star})\right\|_2^2
    \right],
\end{equation}
computed only over valid and present keypoints. To handle trace truncation, a validity head pools the future control-point tokens per keypoint and predicts an $H$-dimensional per-step validity logit, trained via sigmoid cross-entropy $\mathcal{L}_{\text{done}}$,

\begin{equation}
    \mathcal{L}_{\text{done}}
    = \frac{\sum_{t=1}^{H} \ell_{\text{BCE}}(\hat{d}_{n,t}, y_{n,t})}
            {N},
    \label{eq:done_loss}
\end{equation}

where $\hat d_{n,t}$ is the predicted per-step validity logit for keypoint $n$ at future
step $t$, $y_{n,t}\in\{0,1\}$ is the ground-truth per-step validity. At inference, this head provides a stop index to freeze the decoded trace past its predicted end.

We also introduce an auxiliary rigidity loss $\mathcal{L}_{\text{rig}}$ to preserve spatial structural consistency by regularizing the clean control points reconstructed in-flight, $\hat{\mathbf{P}}_n=\mathbf{P}_{n}^{\tau}-\tau v_{\theta}$. Inspired by \citet{liu2025trace}, this loss penalizes non-rigid deformations within the same object part. Unlike prior work that relies on ground-truth object segmentation masks available only in synthetic environments, we use the DINO cluster identities produced by \datapipe. Within each cluster, the pairwise distance between control points of different keypoints should remain invariant across the control-point sequence:
\begin{equation}
    \mathcal{L}_{\text{rig}}
    = \mathbb{E}_{\tau,\boldsymbol{\epsilon}}
    \left[
    \frac{1}{|R|}\sum_{(n,n')\in R}
    \operatorname{Var}_{d}\left(
    \left\|\hat{\mathbf{P}}_{n,d}-\hat{\mathbf{P}}_{n',d}\right\|_2^2
    \right)
    \right],
\end{equation}
where $R$ is the set of unique keypoint pairs sharing a part cluster identity and $d\in\{1,\dots,D\}$ indexes control points. The joint objective is
\begin{equation}
    \mathcal{L}
    = \mathcal{L}_{\text{flow}}
    + \lambda_{\text{done}}\mathcal{L}_{\text{done}}
    + \lambda_{\text{rig}}\mathcal{L}_{\text{rig}}.
\end{equation}
At inference, we integrate $v_{\theta}$ with a 4-step Euler scheme on $\tau\in[1,0]$ and decode the absolute trace through the B-spline basis.

\subsection{Trace-Conditioned Action Expert Details}
\label{subsec:appendix_action_expert}

The action expert conditions on intermediate trace features rather than fully denoised traces. Following partial-denoising schemes used in recent work~\citep{hu2025video, wang2026lamp}, we initialize a pure-noise control-point input $\mathbf{P}^{1}\sim\mathcal{N}(\mathbf{0},\mathbf{I})$ and simulate a single step of the 4-step Euler solver. We then extract the intermediate hidden states of the Trace Expert as the motion descriptor $\mathbf{z}_{\text{trace}}$. This single step preserves task-relevant dynamics while avoiding the computational cost of a full rollout.

To inject 3D dynamics without disrupting pretrained VLM representations, we fuse $\mathbf{z}_{\text{trace}}$ into the last-layer VLM features through a gated cross-attention module. Let $\tilde{\mathbf{h}}_{\text{trace}}=\mathrm{LN}(\mathbf{W}_{\text{proj}}\mathbf{z}_{\text{trace}})$ denote the projected motion features. The guided features are
\begin{equation}
    \mathbf{z}_{\text{guided}}
    = \mathbf{z}+\sigma(g)\cdot
    \mathrm{CA}\!\left(Q=\mathrm{LN}(\mathbf{z}),\;K=V=\tilde{\mathbf{h}}_{\text{trace}}\right),
\end{equation}
where $\mathrm{CA}$ denotes multi-head cross-attention, $\mathbf{z}$ denotes the last-layer VLM features, and $g$ is a learnable scalar gate shared across all heads and spatial positions. We initialize $g$ at zero and pass it through a sigmoid $\sigma(\cdot)$ to bound the gate within $(0,1)$, starting the policy in a weak motion-injection regime that strengthens only when beneficial.

The action expert adopts the self-attention architecture of $\pi_{0.5}$~\citep{intelligence2025pi_} and generates continuous actions via flow matching. Beyond the guided features $\mathbf{z}_{\text{guided}}$, which serve as the conditioning prefix, the expert receives three additional inputs and tokenizes each through a dedicated stem: a gripper-camera image encoded by DINOv2~\citep{oquab2023dinov2}, robot proprioception mapped through an MLP, and the language instruction. The noisy action sequence enters the expert as the query. We define a linear action path $\mathbf{a}^{\tau}=(1-\tau)\boldsymbol{\epsilon}_a+\tau\mathbf{a}$ with $\boldsymbol{\epsilon}_a\sim\mathcal{N}(\mathbf{0},\mathbf{I})$, and train the velocity field $v_{\phi}$ with
\begin{equation}
    \mathcal{L}_{\text{action}}
    = \mathbb{E}_{\tau,\mathbf{a},\boldsymbol{\epsilon}_a}
    \left\|v_{\phi}\!\left(\mathbf{a}^{\tau},\tau,\mathbf{z}_{\text{guided}},\mathbf{c}\right)
    - (\mathbf{a}-\boldsymbol{\epsilon}_a)\right\|_2^2,
\end{equation}
where $\mathbf{c}$ collects the proprioception, gripper-camera, and language conditions.

\section{Model Training}

\subsection{Training Strategy}
\label{subsec:training_strategy}

To ensure \modelname can robustly predict traces even when historical traces or metric depth are unavailable at inference, we apply a two-level history dropout where we either drop the historical traces of all keypoints simultaneously with probability $0.5$, or drop each keypoint's history independently with probability $0.3$. The metric depth channel is randomly omitted with probability $0.7$ to allow the model to flexibly fall back on static RGB observations. The pretrained VLM backbone is kept frozen to preserve its generalist visual-language representations, allowing only the trace expert and projection layers to adapt to 3D kinematics.

\section{Experiment Details}
\subsection{Metric}\label{sec:appendix_metrics}

For each evaluation slot, the model predicts $16$ keypoint traces over $T=\{8, 16, 32\}$
future timesteps in normalized image--depth space
$(u, v, z) \in [-1, 1]^2 \times \mathbb{R}_{\geq 0}$
(UV in the resized $256{\times}256$ camera frame, $z$ in metric meters). Let $\hat{\tau}^{(s)}_{k} \in \mathbb{R}^{T \times 3}$
denote the $s$-th sample for keypoint $k$, and $\tau^{\star}_{k}$ its ground-truth
future. All distances below use the Euclidean pointwise cost in $(u, v, z)$ space
and are reported as means over valid future steps.

\begin{itemize}[leftmargin=2.0em,labelsep=1.2em,itemsep=0.2em,topsep=0in]
    \item \textbf{minADE and minFDE.} Average Displacement Error (ADE) and Final Displacement Error (FDE) compute the mean Euclidean distance over all predicted timesteps and the final timestep, respectively. We take the minimum over the $S$ samples:
    $$
        \mathrm{minADE} \;=\; \mathbb{E}_{\text{slot}}\!\left[ \frac{1}{K}\sum_{k=1}^{K} \min_{s \in [S]} \frac{1}{T} \sum_{t=1}^{T} \|\hat{\tau}^{(s)}_{k,t} - \tau^{\star}_{k,t}\|_2 \right],
    $$
    $$
        \mathrm{minFDE} \;=\; \mathbb{E}_{\text{slot}}\!\left[ \frac{1}{K}\sum_{k=1}^{K} \min_{s \in [S]} \|\hat{\tau}^{(s)}_{k,T} - \tau^{\star}_{k,T}\|_2 \right].
    $$

    \item \textbf{minDTW.} For each keypoint we compute the Dynamic Time Warping distance between each sample and the GT, and take the minimum
    over samples:
    $$
        \mathrm{minDTW}
        \;=\;
        \mathbb{E}_{\text{slot}}\!\left[
            \frac{1}{K}\sum_{k=1}^{K}
            \min_{s \in [S]}
            \mathrm{DTW}\!\left(\hat{\tau}^{(s)}_{k},\, \tau^{\star}_{k}\right)
        \right].
    $$
    DTW allows monotonic time warping, so it scores the \emph{shape} of the
    predicted path independent of small temporal misalignments.

    \item \textbf{minFD.} Identical aggregation, with the discrete Fréchet
    distance replacing DTW:
    $$
        \mathrm{minFD}
        \;=\;
        \mathbb{E}_{\text{slot}}\!\left[
            \frac{1}{K}\sum_{k=1}^{K}
            \min_{s \in [S]}
            \mathrm{FD}\!\left(\hat{\tau}^{(s)}_{k},\, \tau^{\star}_{k}\right)
        \right].
    $$
    Whereas DTW averages pointwise displacement after alignment, the Fréchet
    distance is the \emph{maximum} pointwise displacement over the optimal
    monotonic alignment, so it is sensitive to large excursions and endpoint
    errors that DTW averages away. We report it alongside model parameters in this appendix, as downstream consumers care about worst-case deviations along the path.

    \item \textbf{Inference time and Parameters.} Inference time represents the mean wall-clock latency per slot on a single A6000 GPU (reported in the main text). Total parameter counts for the respective models are provided alongside the FD results.
\end{itemize}

\subsection{Trace-Prediction Evaluation Set}
\label{sec:appendix_eval_set}

The evaluation set introduced in Sec.~\ref{subsec:trace_generation} is held out from the \datapipe pretraining corpus at the source/environment level, so none of its videos are seen during \modelname pretraining. It comprises Human (30 clips), Mobile ALOHA (30), Franka (30), UR3 (30), and VIMA-Bench (9) demonstrations, for 129 total evaluation slots.

We separately quantify how much the B-spline fit itself alters the training target: the reconstruction error between B-spline-fitted and raw traces averages $\sim$0.02 DTW across $T\in\{8,16,32\}$.

Approximately 28\% of the evaluation slots involve bimanual human or robot interactions (e.g., coordinated two-arm manipulation and hand-object contact), included in the aggregate metrics of Table~\ref{tab:ade_fde_dtw_results} rather than broken out separately.

\begin{table*}[h!]
\centering
\caption{\textbf{2D and 3D trace prediction evaluation (Fréchet Distance and parameters)}. Comparison over time horizons $T \in \{8, 16, 32\}$. All baselines receive the same image and text pairs, except $^{\ddagger}$ which requires depth input.}
\label{tab:fd_params_results}
\resizebox{0.75\textwidth}{!}{%
\begin{tabular}{cl ccc ccc c}
\toprule
& \multirow{2}{*}{\textbf{Method}} & \multicolumn{3}{c}{\textbf{top1-FD $\downarrow$}} & \multicolumn{3}{c}{\textbf{top5-FD $\downarrow$}} & \multirow{2}{*}{\textbf{Params}} \\
\cmidrule(lr){3-5} \cmidrule(lr){6-8}
& & $T=8$ & $16$ & $32$ & $8$ & $16$ & $32$ & \\
\midrule
\multirow{6}{*}{\textbf{2D}} 
& Gemini 3.1 Pro & 0.324 & 0.467 & 0.505 & 0.269 & 0.385 & 0.416 & N/A \\
& Gemini 3 Flash & 0.324 & 0.467 & \textbf{0.504} & 0.266 & 0.387 & 0.417 & N/A \\
& GPT-5.5        & 0.342 & 0.476 & 0.511 & 0.299 & 0.415 & 0.449 & N/A \\
& Track2Act \cite{bharadhwaj2024track2act}      & 0.363 & 0.543 & 0.631 & 0.304 & 0.420 & 0.451 & 0.47B \\
& Hamster \cite{li2025hamster}       & 0.339 & 0.462 & 0.505 & 0.291 & 0.390 & 0.429 & 13.5B \\
& \textbf{\modelname (Ours)} & \textbf{0.314} & \textbf{0.446} & 0.517 & \textbf{0.200} & \textbf{0.306} & \textbf{0.370} & 2.59B \\
\midrule
\multirow{5}{*}{\textbf{3D}} 
& 3DFlowAction \cite{zhi20253dflowaction}   & 0.765 & 0.843 & 0.866 & 0.664 & 0.747 & 0.772 & 2.04B \\
& Dream2Flow$^{\ddagger}$ \cite{dharmarajan2026dream2flow}     & 0.547 & 0.710 & 0.787 & 0.325 & 0.464 & 0.530 & 11.3B \\
& TraceGen$^{\ddagger}$ \cite{lee2026tracegen}    & 0.450 & 0.560 & 0.642 & 0.291 & 0.395 & 0.457 & 0.67B \\
& \textbf{\modelname(Ours)}  & \textbf{0.329} & \textbf{0.455} & \textbf{0.527} & \textbf{0.210} & \textbf{0.319} & \textbf{0.384} & 2.59B \\ 
\bottomrule
\end{tabular}%
}
\end{table*}

\subsection{Additional results on Trace Prediction}\label{sec:appendix_trace_qual}
\noindent\textbf{Fréchet distance comparison.}
Beyond the metrics in the main paper, we further evaluate trace prediction quality using the Fréchet distance (FD), which measures the geometric similarity between predicted and ground-truth traces while accounting for their ordering along the path. As shown in Table~\ref{tab:fd_params_results}, we report Top-1 FD and Top-5 FD across time horizons $T \in \{8, 16, 32\}$ against both 2D and 3D baselines, where all methods receive the same image–text pairs except for depth-conditioned baselines. Our method achieves the strongest overall FD performance for both metrics, indicating that our predicted traces are not only accurate on average but also geometrically faithful to the ground-truth motion, and that this advantage holds consistently as the prediction horizon grows.

\noindent\textbf{Parameter efficiency.}
Table~\ref{tab:fd_params_results} also reports performance relative to model size. Our method attains strong trace prediction performance while maintaining a favorable parameter-efficiency trade-off compared with competing baselines. For a fair comparison, we report the parameter count of each baseline at trace inference time. Specifically, we count every component that participates in the forward pass producing the predicted trace, including frozen pretrained backbones, diffusion U-Nets, vision encoders, and trace prediction heads. Components used only during training, such as teacher networks or auxiliary heads, and components used only in downstream action execution, such as separate residual policies or optimization-based action solvers, are excluded. For methods with released checkpoints, parameter counts are obtained by summing \texttt{numel()} over all loaded parameters. For closed-source models or methods without public checkpoints, we use reported model sizes when available and otherwise mark the parameter count as undisclosed.

\begin{figure*}[h!]
    \centering
    \includegraphics[width=\textwidth]{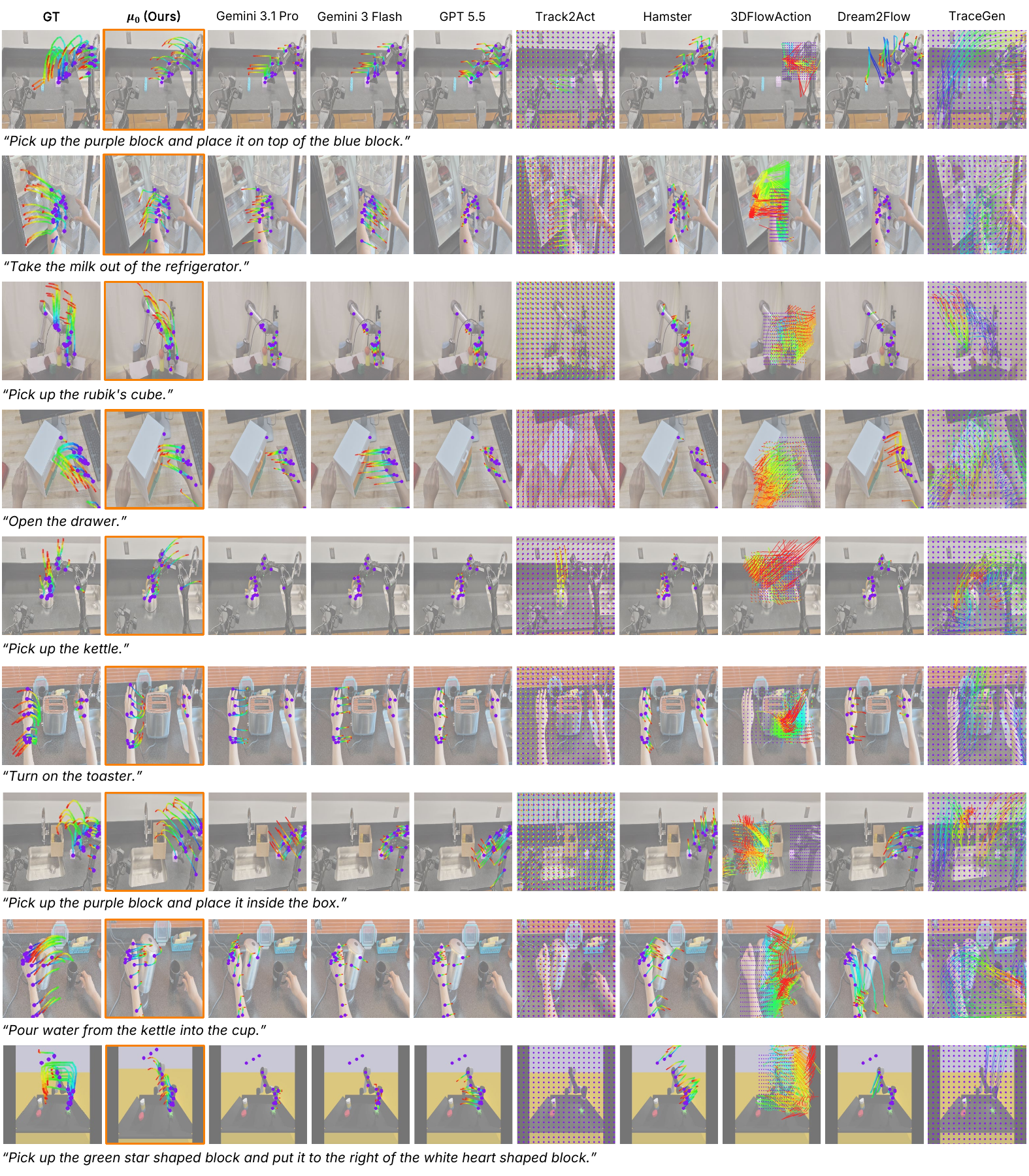}
    \vspace{-1em}
    \caption{\textbf{Additional qualitative comparisons.} We show predicted traces across all methods on additional manipulation tasks, one per row, with the language instruction shown below each example. Columns follow the same ordering as the main paper: ground truth (GT), our method (\modelname), general-purpose VLMs (Gemini 3.1 Pro, Gemini 3 Flash, GPT 5.5), and trace prediction baselines (Track2Act, Hamster, 3DFlowAction, Dream2Flow, TraceGen). Across diverse tasks, our method consistently produces coherent, task-relevant traces that better align with the intended manipulation dynamics, while baselines often generate sparse, noisy, overly dense, or spatially misaligned traces.}
    \label{fig:qual_appendix}
\end{figure*}

\noindent\textbf{Qualitative results.}
Figure~\ref{fig:qual_appendix} presents additional qualitative comparisons across a diverse set of manipulation tasks. Consistent with the quantitative results above, \modelname consistently produces coherent, task-relevant traces that better align with the intended manipulation dynamics, whereas baselines often generate sparse, noisy, overly dense, or spatially misaligned traces.

\subsection{Deployment Cost: Parameters and Latency}
\label{sec:appendix_deployment_cost}

Table~\ref{tab:param_breakdown} reports the full inference-time parameter breakdown for \modelname{}+action expert alongside $\pi_0$ and $\pi_{0.5}$, as used in the RoboCasa365 and real-world experiments (Sec.~\ref{subsec:action_generation}). $\pi_0$ and $\pi_{0.5}$ share an identically sized 3.45B-parameter backbone. $\pi_0$'s backbone is pretrained on 903M action-labeled frames, 69.5$\times$ larger than our 13M-frame, action-free \datapipe corpus; the pretraining scale of $\pi_{0.5}$'s backbone is not publicly disclosed. Each baseline then adds a task-specific trainable head (578.0M parameters for $\pi_0$, 693.4M for $\pi_{0.5}$). \modelname{}+action expert instead pairs a 2.59B frozen trace backbone (Sec.~\ref{sec:training}) with a 457.0M-parameter trainable head (436.4M action expert plus a 20.6M feature mixer), for 3.04B parameters total---fewer than either baseline.

\begin{table}[h]
\centering
\caption{\textbf{Inference-time parameter breakdown.} $\pi_0$ and $\pi_{0.5}$ are fine-tuned from public pretrained checkpoints; all methods share the same demonstrations, training steps, and batch size (Table~\ref{tab:robocasa_training_hyperparameters}).}
\label{tab:param_breakdown}
\begin{tabular}{lccc}
\toprule
\textbf{Model} & \textbf{Frozen Backbone} & \textbf{Trainable (Head)} & \textbf{Total} \\
\midrule
$\pi_0$ & 3.45B & 578.0M & 4.03B \\
$\pi_{0.5}$ & 3.45B & 693.4M & 4.14B \\
\modelname{} + AE (Ours) & \textbf{2.59B} & \textbf{457.0M} (436.4M AE + 20.6M Mixer) & \textbf{3.04B} \\
\bottomrule
\end{tabular}
\end{table}

We also measure end-to-end inference latency, the full wall-clock time from observation to an executable action chunk, including both trace-denoising and the action-expert forward pass. \modelname{}+action expert runs at 195.9--211.2\,ms per action chunk (mean 200.8\,ms), compared to 171.1--178.9\,ms (mean 173.0\,ms) for $\pi_{0.5}$: an increase of approximately 16\%, attributable to the additional trace-denoising stage.

\subsection{RoboCasa365}\label{sec:appendix_robocasa}

\noindent\textbf{Environment details.}
We evaluate simulated action generation in RoboCasa365~\citep{nasiriany2026robocasa365}, a large-scale household manipulation benchmark built on the RoboCasa kitchen simulation platform~\citep{nasiriany2024robocasa}. The benchmark provides diverse kitchen layouts, object assets, and task initializations, making it well suited for testing whether policies generalize across scene and object variation rather than memorizing a fixed setup. We use the PandaOmron mobile manipulator, which consists of a Franka Panda arm mounted on an Omron mobile base and equipped with a gripper. Each policy observes two $256 \times 256$ RGB inputs, a left third-person camera image and a wrist/gripper camera image, together with a language instruction and a 16-dimensional proprioceptive state. The action space is 12-dimensional, including arm motion, gripper control, and mobile-base control.

We evaluate 8 representative atomic tasks from RoboCasa365: \textit{CloseFridge}, \textit{OpenFridge}, \textit{CoffeeServeMug}, \textit{PickPlaceFridgeShelfToDrawer}, \textit{TurnOnMicrowave}, \textit{SlideToasterOvenRack}, \textit{PickPlaceCounterToCabinet}, and \textit{TurnOnToasterOven}. For each task, we use 100 demonstrations, resulting in 800 demonstrations in total. All methods use the same demonstrations, RGB observations, language instructions, and proprioceptive states. TraceGen additionally requires depth input, so we estimate depth from RGB observations using Depth Anything V2~\citep{yang2024depth} and provide the predicted depth images only to TraceGen.

\begin{table}[t!]
\centering
\footnotesize
\caption{\textbf{RoboCasa365 training hyperparameters.} We use the same hyperparameters for all methods.}
\label{tab:robocasa_training_hyperparameters}
\begin{tabular}{@{}lc@{}}
\toprule
Hyperparameter & Value \\
\midrule
Action dimension & 12 \\
Action horizon & 16 \\
Execution horizon & 8 \\
Batch size & 32 \\
Optimizer & AdamW \\
Learning rate & $1 \times 10^{-4}$ \\
Warmup steps & 1{,}000 \\
Training steps & 50{,}000 \\
\bottomrule
\end{tabular}
\end{table}

\begin{figure*}[!h]
    \centering
    \includegraphics[width=\textwidth]{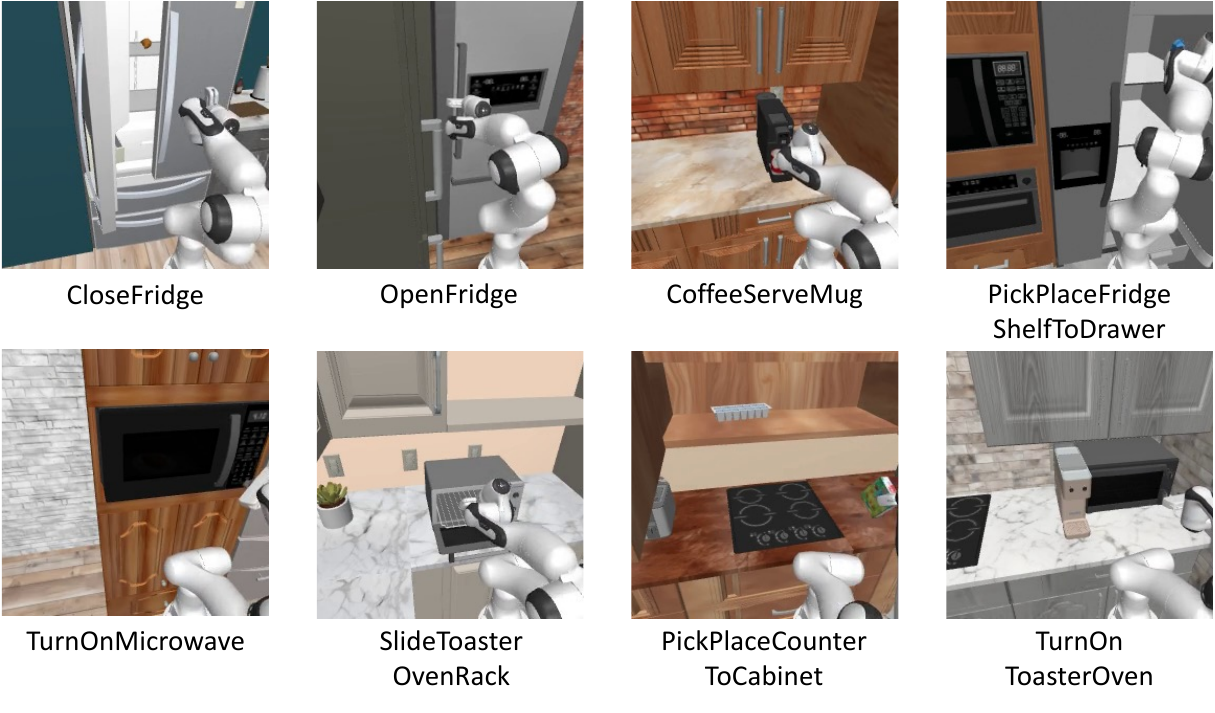}
    \vspace{-1em}
    \caption{\textbf{RoboCasa365 simulation examples.} Example evaluation scenes from the 8 RoboCasa365 tasks used in our simulated experiments. The benchmark randomizes scene layouts, object instances, and initial configurations across rollouts, emphasizing policy generalization rather than memorization of a fixed scene.}
    \label{fig:robocasa_examples}
\end{figure*}

\noindent\textbf{Training details.}
We use the LeRobot~\citep{cadene2026lerobot} implementations of Diffusion Policy~\citep{chi2025diffusion}, $\pi_0$~\citep{black2025pi0}, and $\pi_{0.5}$~\citep{intelligence2025pi_}. Diffusion Policy is trained from scratch on the target RoboCasa365 demonstrations using the multi-task DiT policy. For $\pi_0$, $\pi_{0.5}$, TraceGen + action expert, and \modelname{} + action expert, we freeze the pretrained backbone and train only the action expert on the target demonstrations. Thus, our method uses the frozen \modelname{} trace model as a motion-prior feature extractor while optimizing a RoboCasa365-specific action expert for control. All training runs are performed on 4 NVIDIA L40S GPUs. Table~\ref{tab:robocasa_training_hyperparameters} summarizes the shared training hyperparameters.

We evaluate each trained policy over 50 rollouts per task. During evaluation, RoboCasa365 randomizes scene layouts, object instances, and initial configurations across rollouts. We use the default sparse task-completion signal from the environment and report the success rate (\%) for each task, together with the average success rate across all eight tasks.

\subsection{Long-Horizon Evaluation}
\label{sec:appendix_long_horizon}

The 8 RoboCasa365 tasks evaluated in the main paper are atomic, single-stage tasks. To characterize performance on longer-horizon manipulation, we further evaluate on three RoboCasa tasks decomposed into 2--3 ordered subtasks with scene and object randomization, over 50 rollouts per task, under the same protocol as the main RoboCasa365 experiments.

Table~\ref{tab:long_horizon} reports subtask-level success rates. \modelname{} matches or outperforms $\pi_0$ on 5 of 7 subtasks and trails $\pi_{0.5}$ on most subtasks. All three methods reach 0\% success on the final subtask of \textit{Organize Condiments}.

\begin{table*}[t]
\centering
\small
\caption{\textbf{Long-horizon RoboCasa evaluation.} Subtask-level success rate (\%) over 50 rollouts/task. \textbf{Bold} and \underline{underline} indicate the best and second-best result per column.}
\label{tab:long_horizon}
\resizebox{0.95\textwidth}{!}{
\begin{tabular}{lccccccc}
\toprule
\multirow{2}{*}{\textbf{Model}} & \multicolumn{2}{c}{\textbf{Pastry Display}} & \multicolumn{2}{c}{\textbf{Place Ice In Cup}} & \multicolumn{3}{c}{\textbf{Organize Condiments}} \\
 & Subtask 1 & Subtask 2 & Subtask 1 & Subtask 2 & Subtask 1 & Subtask 2 & Subtask 3 \\
\midrule
$\pi_0$ & 40 & 2 & 22 & 2 & \underline{28} & \textbf{8} & 0 \\
$\pi_{0.5}$ & \textbf{44} & \textbf{6} & \textbf{32} & \textbf{6} & \textbf{32} & \textbf{8} & 0 \\
\modelname{} + AE (Ours) & \textbf{44} & \underline{4} & \underline{26} & \underline{4} & 22 & 6 & 0 \\
\bottomrule
\end{tabular}
}
\end{table*}

\subsection{Real-world Robot}\label{sec:appendix_real_robot}

\noindent\textbf{Hardware setup.}
Figure~\ref{fig:real_robot_setup} shows the real-robot platforms that we use for both demonstration collection and policy evaluation. A fixed-base UR3 manipulator with a two-finger gripper executes all manipulation tasks. Two RGB cameras, mounted respectively at a third-person viewpoint and on the wrist, provide $224 \times 224$ visual observations. The robot proprioception comprises the 6D end-effector pose and the gripper state, yielding a 7D state vector. A human teleoperator collects demonstrations by controlling the end-effector pose and gripper command through a custom teleoperation interface.

\noindent\textbf{Training details.}
Table~\ref{tab:real_robot_training_hyperparameters} reports the task-specific training hyperparameters, which we share across all methods to ensure a fair comparison. For our method, we freeze the VLM backbone and the trace expert of \modelname, and we train only the action expert from scratch on the collected demonstrations.

\begin{table*}[t!]
\centering
\footnotesize
\caption{\textbf{Real-robot training hyperparameters.} We use the same task-specific hyperparameters for all methods. Each subtable corresponds to one real-world task.}
\label{tab:real_robot_training_hyperparameters}
\begin{minipage}[t]{0.31\textwidth}
\centering
\textbf{Pick \textless object\textgreater{} into Sink}

\vspace{0.25em}
\begin{tabular}{@{}lc@{}}
\toprule
Hyperparameter & Value \\
\midrule
Action dimension & 7 \\
Action horizon & 50 \\
Execution horizon & 25 \\
Batch size & 32 \\
Optimizer & AdamW \\
Learning rate & $5 \times 10^{-5}$ \\
Warmup steps & 400 \\
Training steps & 8{,}000 \\
\bottomrule
\end{tabular}
\end{minipage}
\hfill
\begin{minipage}[t]{0.31\textwidth}
\centering
\textbf{Pour Almonds into \textless object\textgreater{}}

\vspace{0.25em}
\begin{tabular}{@{}lc@{}}
\toprule
Hyperparameter & Value \\
\midrule
Action dimension & 7 \\
Action horizon & 50 \\
Execution horizon & 25 \\
Batch size & 32 \\
Optimizer & AdamW \\
Learning rate & $5 \times 10^{-5}$ \\
Warmup steps & 300 \\
Training steps & 6{,}000 \\
\bottomrule
\end{tabular}
\end{minipage}
\hfill
\begin{minipage}[t]{0.31\textwidth}
\centering
\textbf{Unfold Towel}

\vspace{0.25em}
\begin{tabular}{@{}lc@{}}
\toprule
Hyperparameter & Value \\
\midrule
Action dimension & 7 \\
Action horizon & 50 \\
Execution horizon & 25 \\
Batch size & 32 \\
Optimizer & AdamW \\
Learning rate & $5 \times 10^{-5}$ \\
Warmup steps & 300 \\
Training steps & 6{,}000 \\
\bottomrule
\end{tabular}
\end{minipage}
\end{table*}

\section{Additional Results}

\subsection{Trace Prediction Quality Ablations}
\label{subsec:ablation_trace}

To validate the architectural and optimization design choices in \modelname, we evaluate the impact of core components on trace prediction quality by systematically disabling or modifying them.

\textbf{Architectural and Optimization Design.}
We first isolate the contributions of our specific modeling choices:
\begin{itemize}[leftmargin=2.0em,labelsep=1.2em,itemsep=0.2em,topsep=0in]
    \item \textbf{w/o B-spline parameterization:} Instead of predicting $D=10$ B-spline control points, the model directly regresses the raw $H=32$ anchor-relative future steps.
    \item \textbf{w/o DINOv2 features:} We remove the per-keypoint patch-feature injection (Eq.~\ref{eq:dino_fuse_appendix}), forcing the model to rely solely on the global vision--language prefix without explicit part-level semantics.
    \item \textbf{Rigidity loss variations:} We experiment with modifying the weight of the auxiliary rigidity loss ($\lambda_{\text{rig}}$) and completely removing it ($\lambda_{\text{rig}}=0$) to measure its impact on preserving intra-part physical consistency.
\end{itemize}

\textbf{Input Modality Robustness.}
Furthermore, we analyze the model's robustness to missing or degraded input modalities. Specifically, we train variants where metric depth is omitted (\textbf{w/o Depth}) and where the short past trajectory is removed (\textbf{w/o Historical Trace}), forcing the model to predict future motion from a static RGB observation alone.

\subsection{Downstream Policy Performance Ablations}
\label{subsec:ablation_action}

We further ablate two design choices in the trace-conditioned action expert (Sec.~\ref{subsec:action_expert}): whether to condition on intermediate trace-denoising features versus the fully decoded trace, and whether the cross-attention gate $\sigma(g)$ should be learned or fixed. Both variants are evaluated on RoboCasa365 with the same 400M-parameter action head used in the main paper (Table~\ref{tab:robocasa_results}).

\noindent\textbf{Full decoded trace vs.\ intermediate features.} Conditioning the action expert on the fully decoded 3D trace---running the trace expert to completion and feeding the resulting coordinates, rather than reading a single partial-denoising step (Appendix~\ref{subsec:appendix_action_expert})---reduces average success from 30.25\% to 29.5\% (Table~\ref{tab:action_expert_ablation}).

\noindent\textbf{Learnable vs.\ fixed gate.} Fixing the cross-attention gate at $\sigma(g){=}0.5$ or $\sigma(g){=}0.25$, instead of learning it, reduces average success to 26.0\% and 25.0\% respectively, underperforming the learnable gate. Across training, the learned gate converges to $\sigma(g)\approx0.48$--$0.50$ on both RoboCasa365 (two seeds: 0.483, 0.481) and all three real-world tasks (Pick-and-Place: 0.499, Pour: 0.500, Unfold Towel: 0.499).

\begin{table*}[t]
    \centering
    \begin{minipage}[t]{0.48\textwidth}
        \centering
        \caption{\textbf{Trace prediction quality ablations.} Trace prediction quality when varying architectural, optimization, and input-modality choices.}
        \resizebox{0.99\linewidth}{!}{%
\begin{tabular}{l ccc}
\toprule
& \multicolumn{3}{c}{\textbf{top5-DTW $\downarrow$}} \\
\cmidrule(lr){2-4}
\textbf{Model Variant} & \textbf{$T=8$} & \textbf{$16$} & \textbf{$32$} \\
\midrule
\multicolumn{4}{l}{\textit{Architecture Variations}} \\
\textbf{Full \modelname} & \textbf{0.127} & \textbf{0.187} & \textbf{0.223} \\
w/o B-spline (Raw Trace) & 0.156 & 0.222 & 0.258 \\
w/o DINOv2 features      & 0.139 & 0.193 & 0.230 \\
w/o Rigidity Loss        & 0.138 & 0.193 & 0.227 \\
\midrule
\multicolumn{4}{l}{\textit{Input Robustness}} \\
w/ Depth $\&$ Trace history     & 0.107 & 0.160 & 0.203 \\
w/o Depth         & 0.112 & 0.168 & 0.207 \\
w/o Trace history            & 0.126 & 0.183 & 0.224 \\
w/o Depth $\&$ Trace history  & 0.127 & 0.187 & 0.223 \\
\bottomrule
\end{tabular}%
}

        \label{tab:ablation}
    \end{minipage}\hfill
    \begin{minipage}[t]{0.48\textwidth}
        \centering
        \caption{\textbf{Downstream policy performance ablations.} Average RoboCasa365 success rate (\%) when conditioning on the fully decoded trace instead of intermediate features, or fixing the cross-attention gate instead of learning it.}
        \small
\renewcommand{\arraystretch}{1.3}
\begin{tabular}{lc}
\toprule
\textbf{Model Variant} & \textbf{Success Rate (\%) $\uparrow$} \\
\midrule
\modelname{} + action expert (Ours) & \textbf{30.25} \\
w/ Full decoded trace & 29.5 \\
\midrule
Fixed gate $\sigma(g){=}0.5$ & 26.0 \\
Fixed gate $\sigma(g){=}0.25$ & 25.0 \\
\bottomrule
\end{tabular}

        \label{tab:action_expert_ablation}
    \end{minipage}
\end{table*}

\subsection{Scaling Analysis}
\label{subsec:scaling_analysis}

A critical property of an effective world model is its ability to scale predictably with increased model capacity and training data. We evaluate this property with three controlled studies. For \textbf{model scaling}, we keep the pretraining dataset fixed and vary model capacity from 342M to 568M and 2.59B parameters. For \textbf{data scaling}, we fix the 2.59B model and train on 5\%, 20\%, and 100\% of the \datapipe dataset. For \textbf{action-head scaling}, we train downstream action heads at two capacities and compare policies with and without frozen trace features. Table~\ref{tab:scaling} reports the full model and data scaling results, and Table~\ref{tab:action_head_scaling} reports the action-head scaling comparison.

\noindent\textbf{Data scaling.}
When the model size is fixed at 2.59B parameters, increasing the pretraining set from 5\% to 100\% improves top5-DTW from 0.134/0.200/0.235 to 0.127/0.187/0.223 for $T{=}8/16/32$. The gains are most consistent at longer horizons, where more diverse interaction videos help the model predict temporally extended motion rather than only short-term displacement.

\noindent\textbf{Model scaling.}
With the full dataset fixed, larger models improve trace prediction across all horizons: the 342M model obtains 0.143/0.205/0.240, the 568M model improves to 0.136/0.191/0.227, and the 2.59B model reaches 0.127/0.187/0.223. This monotonic trend indicates that the trace-prediction objective remains capacity-limited at our current scale.

\noindent\textbf{Action-head scaling.}
Table~\ref{tab:action_head_scaling} shows that frozen trace features improve downstream policy learning for both action-head sizes. With a 200M action head, using trace features raises success from 10.625\% to 25.625\%; with a 400M action head, the gain narrows from 28.25\% to 30.25\%. The larger gap at the smaller 200M head suggests that limited action-head capacity benefits most from trace-space features that provide structured motion information, whereas a larger head can recover much of this signal on its own.

\begin{table*}[t]
    \centering
    \begin{minipage}[t]{0.48\textwidth}
        \centering
        \caption{\textbf{Scaling analysis.} Evaluating the performance impact when scaling model parameters and training data volume.}

\small
\setlength{\tabcolsep}{8pt}
\renewcommand{\arraystretch}{1.15}
\begin{tabular}{l ccc}
\toprule
 & \multicolumn{3}{c}{\textbf{top5-DTW $\downarrow$}} \\
\cmidrule(lr){2-4}
\textbf{Scale Factor} & $\mathbf{T{=}8}$ & \textbf{16} & \textbf{32} \\
\midrule
\multicolumn{4}{@{}l}{\textit{Model Scaling (100\% Data)}} \\
\quad 342M Model  & 0.143 & 0.205 & 0.240 \\
\quad 568M Model  & 0.136 & 0.191 & 0.227 \\
\quad 2.59B Model & \textbf{0.127} & \textbf{0.187} & \textbf{0.223} \\
\midrule
\multicolumn{4}{@{}l}{\textit{Data Scaling (2.59B Model)}} \\
\quad 5\% Dataset   & 0.134 & 0.200 & 0.235 \\
\quad 20\% Dataset  & 0.138 & 0.195 & 0.227 \\
\quad 100\% Dataset & \textbf{0.127} & \textbf{0.187} & \textbf{0.223} \\
\bottomrule
\end{tabular}

        \label{tab:scaling}
    \end{minipage}\hfill
    \begin{minipage}[t]{0.48\textwidth}
        \centering
        \caption{\textbf{Action head scaling.} \modelname with an action expert remains robust across action-head capacities compared to \textit{w/o Trace}.}

\small
\renewcommand{\arraystretch}{1.3}
\begin{tabular*}{\linewidth}{@{\extracolsep{\fill}} l c@{}}
\toprule
\textbf{Model Variant} & \textbf{\begin{tabular}{@{}c@{}}Success \\ Rate (\%) $\uparrow$\end{tabular}} \\
\midrule
\multicolumn{2}{@{}l}{\textit{200M Action Head}} \\
\quad w/o Trace & 10.625 \\
\quad \modelname{} + action expert (Ours) & \textbf{25.625} \\
\midrule
\multicolumn{2}{@{}l}{\textit{400M Action Head}} \\
\quad w/o Trace & 28.25 \\
\quad \modelname{} + action expert (Ours) & \textbf{30.25} \\
\bottomrule
\end{tabular*}

        \label{tab:action_head_scaling}
    \end{minipage}
\end{table*}

\subsection{Robustness to Perception and Trace-Prediction Errors}
\label{sec:appendix_robustness}

Since \modelname is trained entirely on automatically extracted \datapipe supervision, its trace predictions depend on the quality of the underlying captioning and keypoint-tracking pipeline. We evaluate sensitivity to two forms of corrupted input, then test whether errors in \modelname's own trace predictions propagate into the downstream policy.

\noindent\textbf{Caption corruption.} We corrupt the language instruction by randomly dropping or replacing 41\% of its words at test time and re-run trace prediction with no other changes. ADE, FDE, DTW, and Fr\'echet distance increase by 2.9--3.3\% relative to clean captions, averaged across horizons $T\in\{8,16,32\}$.

\noindent\textbf{Keypoint corruption.} We corrupt the input query keypoints by randomly dropping 30\% of points and applying spatial jitter to the remainder, simulating upstream tracking failures. The same metrics increase by 1.2--4.1\%.

\noindent\textbf{Robustness of the downstream policy to trace-prediction error.} We inject zero-mean Gaussian noise ($\sigma=0.05$, in the normalized units of Appendix~\ref{sec:appendix_metrics}) into \modelname's predicted 3D traces before they reach the action expert, and re-evaluate on RoboCasa365 (50 rollouts/task, Appendix~\ref{sec:appendix_robocasa}). Average success decreases from 30.25\% to 27.5\%.

\section{Additional Related Work Discussion}
\label{sec:appendix_related_work}

\noindent\textbf{Extended Discussion on Embodiment-Agnostic World Models.}
While the main text briefly outlines the limitations of existing world models, we provide a more granular breakdown here. Pixel-space video models~\citep{wu2024unleashing, guo2026ctrlworld} and world-action models~\citep{li2026causal, ye2026world, ye2026gigaworld} excel at learning broad visual dynamics. However, they expend the majority of their representational capacity on dense appearance details that are often irrelevant to the geometry and contact structure required for manipulation. 

To alleviate this burden, intermediate representations have been explored~\citep{jang2026lace, zhou2025dinowm, hu2025video, gu2024rt, bharadhwaj2024track2act, wen2024any, vecerik2024robotap, kambara2026lilac, zhi20253dflowaction, wang2026lamp}, yet each distinct prior carries inherent trade-offs:
(1) \textit{Latent features} avoid pixel reconstruction but are notoriously difficult to inspect, control, or translate into precise robot motion.
(2) \textit{2D tracks and optical flow} provide a more geometric interface but lack metric depth, often obscuring crucial 3D contacts and spatial object motion.
(3) \textit{Recent 3D flow methods} restore metric structure but typically predict dense flow fields over fixed grids (wasting compute budget on static backgrounds), condition rollouts on labeled actions, or relegate motion to an auxiliary policy prior rather than building a standalone world model. 
\modelname bypasses these issues by making a sparse set of semantic 3D traces the explicit prediction target. This yields a world-model output that is compact, metric, and directly reusable as a motion interface for downstream policies.

\noindent\textbf{Extended Comparison: Trace Representations and TraceGen.}
As noted in the main text, visual motion plans for manipulation generally fall into three families: VLM-based waypoints~\citep{li2025hamster, zhou2025robotracer, yuan2024robopoint, yang2025magma}, post-hoc track extraction from generated video~\citep{ko2024learning, bharadhwaj2025genact, li2025novaflow, dharmarajan2026dream2flow}, and direct track prediction via diffusion or flow matching~\citep{nguyen2026pixel, gao2025flip, lin2026roboflow4d}. While these choices provide useful auxiliary guidance, they are less suited to learning a universal dynamics model because raw action semantics often depend heavily on specific robot kinematics and control frequencies.

TraceGen~\citep{lee2026tracegen} represents the closest prior work to ours, but \modelname introduces fundamental changes to both the supervision pipeline and the model interface. Specifically, TraceGen relies on fixed-grid traces over short clips, necessitates depth information at inference time, and utilizes a hand-designed trace replay mechanism. It does not provide a reusable, query-conditioned 3D world model.
In contrast, \modelname addresses these limitations end-to-end:
First, our data pipeline, \datapipe, replaces fixed grids with semantic interaction keypoints, global 3D tracking, event-level captions, and movement filtering. 
Second, the Trace Expert in \modelname predicts query-conditioned B-spline futures via semantic flow matching. 
Finally, our Action Expert directly consumes frozen trace-denoising features rather than relying on raw trace replay. 
Through this design, actionable 3D traces are elevated from merely an auxiliary visual cue to the central, video-pretrained motion interface that drives  cross-embodiment manipulation.

\end{document}